\documentclass[accepted]{uai2026}

\usepackage[american]{babel}

\usepackage{natbib} 
\usepackage{mathtools} 
\usepackage{siunitx} 
\usepackage{booktabs} 
\usepackage{tikz}

\usepackage{amsmath}
\usepackage{amssymb}
\usepackage{amsthm}
\usepackage{bm}

\usepackage{hyperref}

\usepackage{microtype}
\usepackage{graphicx}
\usepackage{subcaption}
\usepackage{cancel}

\usepackage[most]{tcolorbox}
\usepackage{xcolor}
\definecolor{acadblue}{RGB}{235, 240, 250} 
\tcbset{
    blue_style/.style={
        colback=acadblue,       
        colframe=black,         
        boxrule=0.5pt,          
        arc=1pt,                
        auto outer arc,
        left=10pt, right=10pt, top=5pt, bottom=5pt, 
        enhanced,
        before skip=10pt,       
        after skip=10pt         
    }
}

\usepackage{tikz}
\usepackage{tikz-3dplot}
\usetikzlibrary{positioning,fit,decorations.pathreplacing,calc,patterns,arrows.meta}
\usepackage{pgfplots}
\pgfplotsset{compat=1.17}

\theoremstyle{plain}
\newtheorem{theorem}{Theorem}[section]
\newtheorem{proposition}[theorem]{Proposition}
\newtheorem{lemma}[theorem]{Lemma}

\theoremstyle{definition}
\newtheorem{definition}[theorem]{Definition}
\newtheorem{assumption}[theorem]{Assumption}
\theoremstyle{remark}
\newtheorem{remark}[theorem]{Remark}

\title{Fourier Analysis on the Boolean Hypercube \\ via Hoeffding Functional Decomposition}

\author[1,2]{Baptiste Ferrere\thanks{Corresponding author.
\href{mailto:baptiste.ferrere@edf.fr}{baptiste.ferrere@edf.fr}}}
\author[1]{Nicolas Bousquet}
\author[2,3]{Fabrice Gamboa}
\author[2,3,4]{Jean-Michel Loubes}
\author[1]{Joseph Muré}

\affil[1]{EDF R\&D, SINCLAIR Laboratory, France}

\affil[2]{IMT, Toulouse, France}

\affil[3]{ANITI, Toulouse, France}

\affil[4]{INRIA Regalia Team, Toulouse, France}
  
\begin{document}
\maketitle

\newcommand{\sixfigscale}{0.7} 
\begin{figure*}[t] 
\centering \includegraphics[width=\dimexpr(\sixfigscale\linewidth)/4-1.5pt\relax]{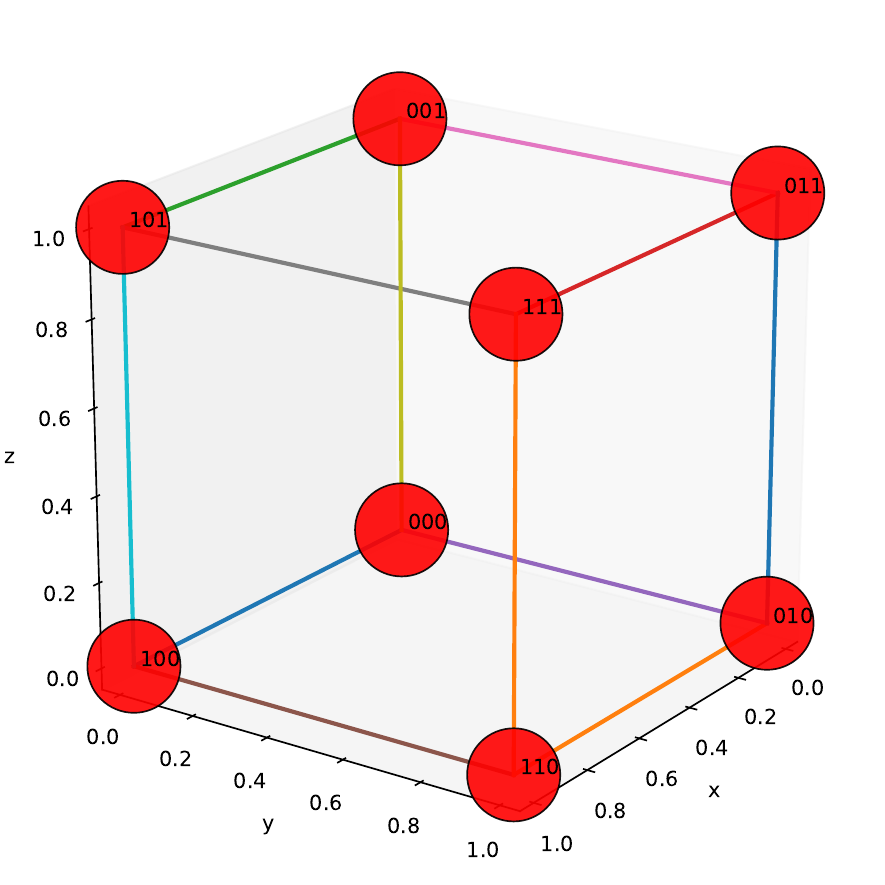}\hfill \includegraphics[width=\dimexpr(\sixfigscale\linewidth)/4-1.5pt\relax]{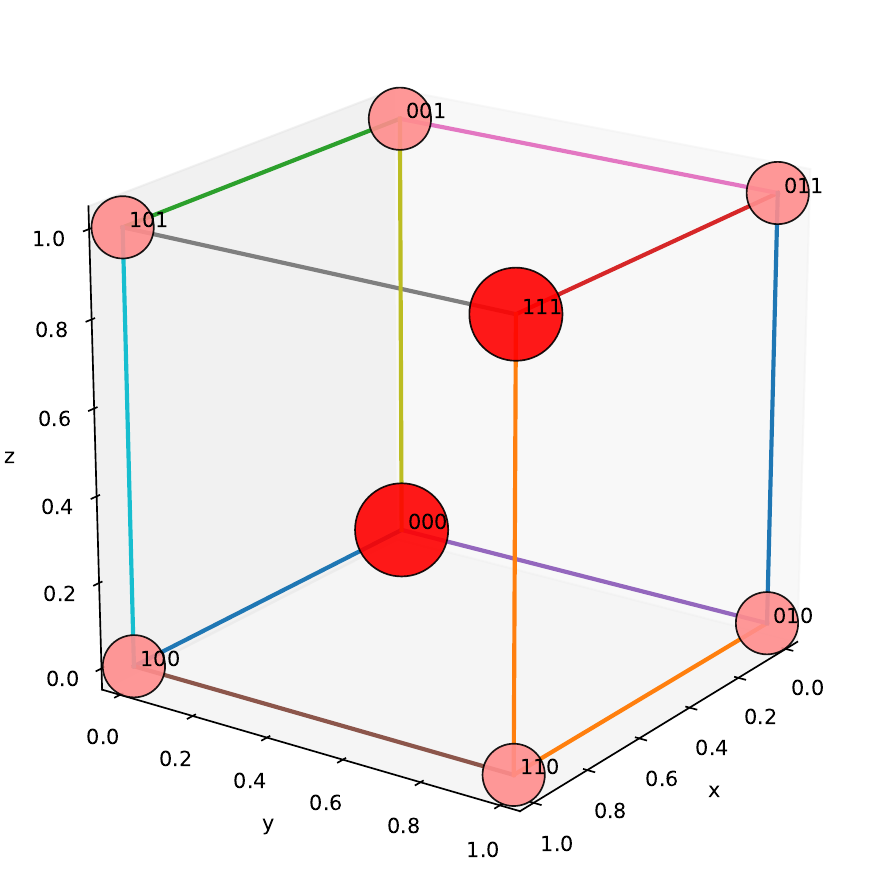}\hfill \includegraphics[width=\dimexpr(\sixfigscale\linewidth)/4-1.5pt\relax]{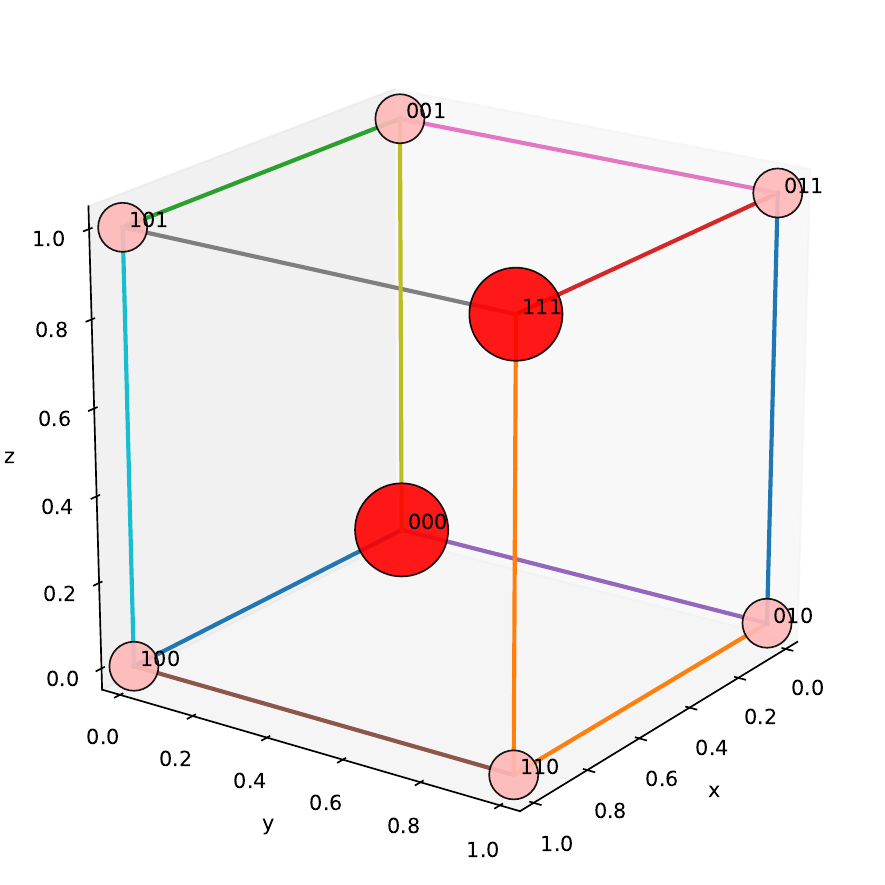}\hfill \includegraphics[width=\dimexpr(\sixfigscale\linewidth)/4-1.5pt\relax]{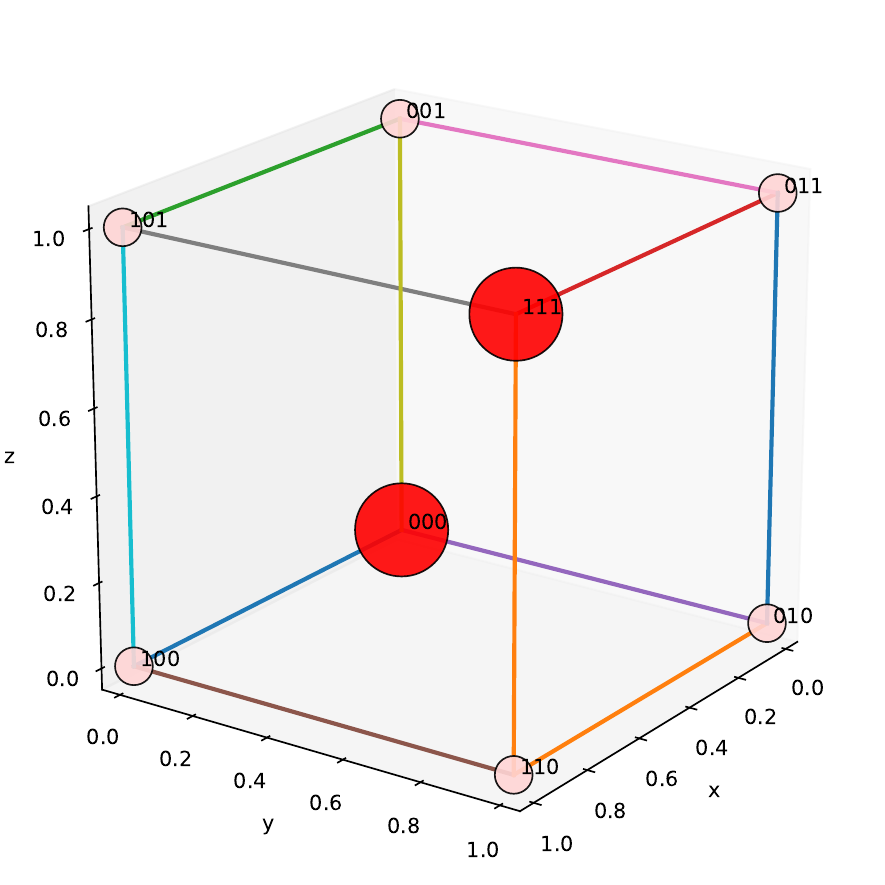}\hfill 
\caption{\textbf{Deformation of the Boolean hypercube.}
On the left, the the mass is uniform over the $2^3$ configurations. As the correlation increases from left to right, probability concentrates on the fully aligned states. At the end, almost all mass is on $000$ and $111$ while all the other configurations vanish.}
\label{fig:hypercubes} 
\end{figure*}

\begin{abstract}
Fourier analysis on the Boolean hypercube is fundamentally defined as the orthogonal decomposition of the space of pseudo-Boolean functions with respect to the uniform probability measure. In this work, we propose an ANOVA-based generalization of the Fourier decomposition on the Boolean hypercube endowed with any arbitrary probability measure. We provide an \emph{explicit} decomposition basis which generalizes the Walsh-Hadamard (or parity functions) basis under any \emph{arbitrary} probability measure on the Boolean hypercube. We formulate the computation of the entire functional decomposition as a least squares problem and also provide a method to address the classical \emph{curse of dimensionality} challenge. We provide a comprehensive generalization of Fourier analysis on the Boolean hypercube, enabling the handling of non-uniform configuration spaces inherent to real-world machine learning tasks, \textit{e.g.} when dealing with \emph{one-hot encoded} features. Finally, we demonstrate its practical impact in the field of explainable AI, by conducting comparative studies with feature attribution methods such as SHAP or TreeHFD.
\end{abstract}

\section{Introduction}\label{sec:intro}
The analysis of \emph{pseudo-Boolean} functions, defined as mappings $f : \{0,1\}^d \to \mathbb{R}$, constitutes a foundational tool in theoretical computer science, canonically formulated via the Boolean Fourier expansion. As detailed in the seminal work of \citet{o2014analysis}, this decomposition enables a precise structural understanding of functions defined on the Boolean hypercube by projecting them onto the \emph{Walsh-Hadamard basis}, composed of the so-called \emph{parity functions}. 

In this paper, we establish that Fourier analysis is a special 
case of the \emph{Hoeffding Functional Decomposition} (HFD) 
\citep{hoeffding_class_1948}---also known as functional ANOVA---and 
leverage this connection to develop a more general framework 
for the analysis of pseudo-Boolean functions. 

Initially introduced to decompose the variance of a function into contributions attributable to individual variables and their interactions, HFD provides a statistically grounded framework to decompose any function $f$ into a sum of functions with variable subsets as arguments. While originally defined for independent inputs, this framework has been rigorously extended to handle dependent input variables, corresponding to non-product measures on the hypercube. A major breakthrough was achieved by \citet{stone1994use} and then \citet{hooker_2007} who generalized the decomposition through an optimization problem under \emph{hierarchical orthogonality} constraints. More recently, \citet{chastaing_generalized_2012} and later \citet{IlIdrissi2023} extended the theoretical validity of this decomposition to broader distributions using \emph{oblique} projections.

This connection is practically motivated: real-world binary data are far from independent. Correlations arise naturally in Ising models \citep{dai2013multivariate}, graphical models \citep{koller2009probabilistic}, genomic data or deterministic constraints such as one-hot encoding, introducing further dependencies. In all these settings, leveraging the HFD framework provides a principled and statistically grounded approach to handle these dependencies.

\paragraph{Contributions.}
In this paper, we address these challenges by providing a framework that bridges Boolean Fourier analysis and the HFD.
\begin{itemize}
    \item \textbf{Closed-form basis decomposition.} We introduce a 
    measure-adaptive basis that extends Boolean Fourier analysis to 
    arbitrary probability measures on $\{0,1\}^d$, yielding a decomposition 
    that is the solution to a natural least-squares problem.

    \item \textbf{Computational tractability.} Despite the exponential complexity of the hypercube, we propose a strategy to overcome the curse of dimensionality. A key practical feature of our framework is that the 
    decomposition is computed once globally, then enables instantaneous 
    local and global explanations across the entire dataset.

    \item \textbf{Connections to XAI.} Experiments on tree ensembles and 
    neural networks show that our approach seamlessly recovers classical 
    importance and interaction patterns, with strong alignment to SHAP 
    \citep{lundberg2017unified, lundberg2018consistent} and the recent 
    TreeHFD algorithm \citep{benard2025tree}.
\end{itemize}

\paragraph{Outline of the paper.}
This paper is organized as follows. Section \ref{sec:related} presents some related work. Section \ref{sec:background} introduces the technical background necessary to develop our methods. In Section \ref{sec:generalized}, we extend the Fourier analysis on the Boolean hypercube based on the generalized HFD framework. Section \ref{sec:numeric} presents comprehensive numerical experiments before concluding in Section \ref{sec:discuss}.

\section{Related Work}\label{sec:related}

The Hoeffding Functional Decomposition (HFD), also known as functional ANOVA/HDMR, expands a square-integrable black-box model \(f(\mathbf X)\) as a sum of components indexed by subsets of variables,
\(f(\mathbf X)=\sum_{S\subseteq[d]} f_S(\mathbf X_S)\), thereby separating main effects and higher-order interactions \citep{hoeffding_class_1948,Rabitz1999}.
When the coordinates of \(\mathbf X\) are independent, the decomposition exists, is unique, and the components are mutually orthogonal, which directly underpins variance-based global sensitivity analysis (GSA) \citep{DaVeiga2021} and Sobol' indices \citep{sobol1990sensitivity}.
For dependent inputs, \citet{stone1994use} and \citet{hooker_2007} clarified that mutual orthogonality is not an imposed identification constraint but rather a consequence of independence, and thus generally fails under realistic correlations \citep{Razavi2021}. To address this, several relaxations have been proposed for general input distributions, including projection-based generalizations and kernel-based constructions \citep{da2015global,chastaing_generalized_2012,IlIdrissi2023}.
From an algorithmic standpoint, estimating HFD components is challenging because it involves conditional expectations under the joint law of \(\mathbf X\); practical estimators often require either strong distributional access or restrictive model classes \citep{lengerich2020purifying}.
Recent progress has made exact or sample-based estimation feasible for tree ensembles \citep{benard2025tree}.

Beyond variance-based indices, cooperative game theory-inspired importance measures allocate a global contribution to each feature by averaging its marginal effect across coalitions, leading to Shapley values and extensions \citep{shapley_notes_1951,harsanyi_1963,owen_shapley_2017,herin2024proportional,LABREUCHE2022225,idrissi2023coalitional}. In GSA, \emph{Shapley effects} were introduced to handle correlated inputs by fairly distributing shared contributions due to both dependence and interaction \citep{song2016shapley}, offering a principled alternative when classical Sobol' indices become hard to interpret under dependence \citep{Razavi2021}. These perspectives emphasize that the choice of the underlying value function is central when inputs are dependent.

Finally, in modern Machine Learning, Shapley values are widely used for local feature attribution; SHAP \citep{lundberg2017unified} popularized this viewpoint and introduced model-agnostic estimators (KernelSHAP) \citep{lundberg2017unified} as well as model-specific algorithms such as TreeSHAP \citep{lundberg2018consistent} for fast \emph{exact} attributions in tree ensembles. Recently, multiple works have established strong connections between Shapley-based explanations and functional decompositions. Indeed \citet{herren2022statistical} relate SHAP estimation to functional ANOVA and highlight the role of the feature distribution; \citet{bordt2023shapley} characterize Shapley values via a family of Shapley interaction explanations and connect them to \emph{ANOVA style} decompositions. For Boolean inputs, very recent work leverages sparse Fourier representations on the Boolean cube to amortize and accelerate SHAP computation in both black-box and tree settings \citep{gorji2024shap}, complementing earlier SHAP estimators by exploiting Fourier analysis.

\begin{figure*}[t]
\centering

\newcommand{\figscale}{1.6} 

\tikzset{
    >=Latex,
    axis/.style={thin, draw=gray!60},
    gridlines/.style={draw=gray!20, very thin},
    origin/.style={circle, fill=black, inner sep=0pt, minimum size=2.2pt},
    v1/.style={->, line width=1.2pt, draw=blue!80!black},  
    v2/.style={->, line width=1.2pt, draw=red!80!black},   
    angarc/.style={dashed, line width=0.8pt, draw=green!60!black},
    anglab/.style={green!60!black, font=\scriptsize}
}

\newcommand{\panelrho}[2]{%
    \def\param{#1}%
    \begin{tikzpicture}[scale=\figscale]
        \pgfmathsetmacro{\angle}{acos(4*\param - 1)}
        \pgfmathsetmacro{\corr}{4*\param - 1} 
        \pgfmathsetmacro{\L}{1}                     
        \pgfmathsetmacro{\r}{0.25}                   
        \pgfmathsetmacro{\angmid}{0.5*\angle}

        \def\marg{0.2} 

        \draw[gridlines, step=0.25] (0,0) grid (1,1);

        \draw[axis, ->] (-0.2,0) -- (1+\marg,0);
        \draw[axis, ->] (0,-0.2) -- (0,1+\marg);

        \foreach \x in {0, 0.5, 1.0} {
            \draw[gray!60] (\x,0) -- (\x,-0.06) node[below, font=\tiny] {\x};
        }
        \foreach \y in {0.5, 1.0} {
            \draw[gray!60] (0,\y) -- (-0.06,\y) node[left, font=\tiny] {\y};
        }

        \node[origin] at (0,0) {};

        \draw[v1] (0,0) -- (\L,0);
        \node[blue!80!black, font=\normalsize] at (\L+0.1, -0.15) {$\psi_{1}$}; 

        \draw[v2] (0,0) -- (\angle:\L);
        \node[red!80!black, font=\normalsize] at (\angle:\L+0.18) {$\psi_{2}$};

        \draw[angarc] (\r,0) arc[start angle=0, end angle=\angle, radius=\r];
        
        \node[anglab] at (\angmid:\r+0.12) {$\theta_{#2}$};

        \node[green!60!black, font=\small] at (0.5, 1.35) {$\mathrm{corr} = \pgfmathprintnumber[fixed, precision=2]{\corr}$};
    \end{tikzpicture}%
}

\begin{minipage}[t]{0.24\textwidth}\centering
    \panelrho{0.25}{1} 
\end{minipage}\hfill
\begin{minipage}[t]{0.24\textwidth}\centering
    \panelrho{0.3125}{2} 
\end{minipage}\hfill
\begin{minipage}[t]{0.24\textwidth}\centering
    \panelrho{0.375}{3} 
\end{minipage}\hfill
\begin{minipage}[t]{0.24\textwidth}\centering
    \panelrho{0.4975}{4} 
\end{minipage}

\caption{Illustration of 2 basis elements in $\mathbb R^2$ across 4 different correlation levels (see Appendix for details).}
\label{fig:GX_dependence_levels}
\end{figure*}

\section{Background}\label{sec:background}

\paragraph{Notations.}
For any integer $d$, we denote by $[d]$ the set $\{1, \dots, d\}$. For any vector $\mathbf x \in \mathbb{R}^d$, we denote by $\mathbf{x}_S$ the subvector of $\mathbf{x}$ whose components are indexed by the set $S \subseteq [d]$. Consequently, $\mathbf{x}_{[d]} = \mathbf x$. Random vectors are denoted with capital letters (\textit{e.g.} $\mathbf{X}$).

\paragraph{Fourier Analysis.}
A key element in the Fourier analysis of the pseudo-Boolean function is the collection of \emph{parity functions} also known under the name of \emph{Walsh-Hadamard} basis that we introduced below.
\begin{definition}
    For any $S \subseteq [d]$ we define the parity function associated to the set $S$ as follows:
    \begin{equation}
        \forall \mathbf x \in \{0,1\}^d, \: \chi_S( \mathbf x) \coloneqq (-1)^{ \sum\limits_{i \in S} \mathbf x_i }
    \end{equation}
\end{definition}
\begin{tcolorbox}[blue_style]
In the analysis of pseudo-Boolean function, we refer to the Fourier representation of $f : \{0,1\}^d \to \mathbb{R}$ as the unique expansion of $f$ given by:
\begin{equation}\label{eq:fourier}
   \forall \mathbf x \in\{0,1\}^d, \: f( \mathbf{x} ) = \sum_{S \subseteq [d]} \widehat{f}(S) \cdot \chi_S(\mathbf{x}),
\end{equation}
where the unique real numbers $\{ \widehat{f}(S)\}_{S \subseteq [d]}$ are called the Fourier coefficients of $f$.
\end{tcolorbox}
In standard Fourier analysis, we consider the inner product between two pseudo boolean functions $f$ and $g$ defined as follows:
\begin{equation}\label{eq:inner_product}
    \left\langle f , g \right\rangle \coloneqq \frac{1}{2^d} \sum\limits_{\mathbf x \in \{0,1\}^d} f( \mathbf x) g( \mathbf x)
\end{equation}
The major consequence of this convention is that the family of parity functions $\left\{ \chi_S \right\}_{ S \subseteq [d] }$ forms an orthonormal basis of the space of pseudo-Boolean functions, \textit{i.e.}
\begin{equation}
\forall S, T \subseteq [d], \quad \left\langle \chi_S , \chi_T \right\rangle = \mathbf{1}_{ \{ S = T \} }.
\end{equation}
Therefore, the Fourier coefficients $\{\widehat{f}(S)\}_{S \subseteq [d]}$ can simply computed via orthogonal projections onto the parity functions:
\begin{equation}
\widehat{f}(S) = \left\langle f, \chi_S \right\rangle = \frac{1}{2^d} \sum\limits_{\mathbf{x} \in \{0,1\}^d}f(\mathbf{x})\chi_S(\mathbf{x}).
\end{equation}
\begin{tcolorbox}[blue_style]
    Finally, the Fourier analysis on the Boolean hypercube can simply be summarized as the orthonormal decomposition of mappings $f : \{0,1\}^d \to \mathbb{R}$. 
\end{tcolorbox}

\paragraph{Distributional Mismatch.}
The Fourier analysis on the Boolean hypercube relies in fact on the prior assumption of a \underline{uniform distribution of binary configurations} over the hypercube. Indeed, if we introduce a random vector $\mathbf X \coloneqq (\mathbf X_1, \dots, \mathbf X_d)$ such that for all $i \in [d], \mathbf X_i$ are i.i.d. $\mathrm{Bernoulli}(1/2)$, the inner product defined previously \eqref{eq:inner_product} strictly equals the following quantity:
\begin{equation}
    \left\langle f, g \right\rangle = \mathbb{E}\left[ f( \mathbf X) g( \mathbf X) \right].
\end{equation}
In other words, the Fourier analysis on the Boolean hypercube is equivalent to assume that the Hilbert space $\mathcal H$ of pseudo-Boolean functions is equipped with the uniform probabilistic measure: any configuration on the Boolean hypercube has the same \emph{weight} of $1/2^d$. In standard machine learning setting, this hypothesis clearly doesn't hold anymore. This mismatch poses a fundamental problem: \textbf{How to adapt the standard Fourier analysis to \emph{any} distribution $\mathbb P$ on the Boolean hypercube?}
We propose linking the decomposition \eqref{eq:fourier} to the Hoeffding Functional Decomposition (HFD) detailed below to answer this question.

\paragraph{HFD via an Optimization Program.}
Here we adopt a more general point of view. Building on the foundational work of \citet{hoeffding_class_1948}, we consider the HFD, formalized as a constrained variational problem by \citet{hooker_2007}. Let $\mathbf{X}$ be a random vector of $\mathbb R^d$. 
\begin{tcolorbox}[blue_style]
We say that a square-integrable function $f(\mathbf{X})$ admits a (generalized) HFD if it can be expanded as the following sum:
\begin{equation}\label{eq:def_anova}
    f(\mathbf{X}) = \sum_{S \subseteq [d]} f_S(\mathbf{X}_S),
\end{equation}
under the \underline{\emph{hierarchical orthogonality condition}}:
\begin{equation}\label{eq:orthogonality}
    \forall T \subsetneq S, \: \forall g \in L^2_T, \quad \mathbb{E}\left[ f_S(\mathbf{X}_S) g(\mathbf{X}_T) \right] = 0,
\end{equation}
where $L^2_T$ denotes the space of square-integrable functions of $\mathbf{X}_T$.
\end{tcolorbox}
This constraint ensures that each term $f_S$ captures interaction effects strictly specific to the subset $S$, orthogonal to any information explainable by lower-order marginals : the represented information decreases as sets $S$ \emph{grow}.

While the HFD provides a powerful framework, the existence and uniqueness of a decomposition satisfying \eqref{eq:orthogonality} are not automatic. They rely on specific assumptions regarding the support and the density of the input distribution. In the general setting, \citet{chastaing_generalized_2012} established that the intersection of the underlying subspaces is well-behaved provided that the probability density is bounded away from zero relative to the product of its marginals.
\begin{assumption}\label{ass:weak_dependence}
    Let $\nu \coloneqq \otimes_{i=1}^d \nu_i$ be a product measure such that $\mathbb{P} \ll \nu$. The density $h \coloneqq d\mathbb{P}/d\nu$ is assumed to satisfy almost surely:
    \begin{equation}\label{eq:assumption_density}
        \exists C > 0, \: \forall S \subseteq [d], \quad h(\mathbf{x}) \geq C \cdot h_S(\mathbf{x}_S) h_{S^c}(\mathbf{x}_{S^c}),
    \end{equation}
    where $h_S$ and $h_{S^c}$ denote the marginal densities of $h$ with respect to $\otimes_{i \in S} \nu_i$ and $\otimes_{i \in S^c} \nu_i$, respectively.
\end{assumption}
\begin{remark}
    The requirement $\mathbb{P} \ll \nu$ implies absolute continuity with respect to a product measure, but it does \emph{not} imply that $\mathbb{P}$ itself is a product measure.
\end{remark}
\begin{theorem}(\citep{chastaing_generalized_2012})
    For any square integrable function $f$ and under Assumption~\ref{ass:weak_dependence}, the generalization of HFD exists and is unique.
\end{theorem}
However, the practical derivation of HFD involves solving a constrained variational problem which is computationally intractable for general dependent inputs, as no closed-form solution exists. This stands in sharp contrast to the independent setting, where the orthogonality condition simplifies significantly, yielding an analytical solution.

\begin{theorem}[Independent HFD \citep{hoeffding_class_1948}]\label{thm:anova_indep}
    If the components of $\mathbf{X}$ are mutually independent, the unique functional ANOVA terms $\{f_S\}_{S \subseteq [d]}$ are given explicitly by the Möbius inversion of conditional expectations:
    \begin{equation}\label{eq:anova_indep}
        f_S(\mathbf{X}_S) = \sum_{ T \subseteq S } (-1)^{ \vert S \vert - \vert T \vert } \mathbb{E}\left[ f(\mathbf{X}) \mid \mathbf{X}_T \right].
    \end{equation}
\end{theorem}
\begin{tcolorbox}[blue_style]
    When $\mathbf{X}_1, \dots, \mathbf{X}_d$ are i.i.d. $\mathrm{Bernoulli}(1/2)$, HFD coincides exactly with standard Fourier analysis:
    \begin{equation}
        \underbrace{ f_S( \mathbf X_S) }_{ \text{HFD component \eqref{eq:def_anova} } } = \underbrace{ \widehat{f}(S) \cdot \chi_S( \mathbf X) }_{ \text{ Fourier component \eqref{eq:fourier} } }
    \end{equation}
\end{tcolorbox}

\paragraph{Problem Formulation.} 
Our goal is to construct a basis of functions $\{ \psi_S \}_{ S \subseteq [d] }$ on the Boolean hypercube such that, for any probability measure $\mathbb{P}$ over $\{0,1\}^d$, the resulting expansion satisfies both the \underline{ANOVA formulation} \eqref{eq:def_anova} and the \underline{hierarchical orthogonality condition} \eqref{eq:orthogonality}. Consistency requires this construction to generalize standard Fourier analysis: specifically, if $\mathbb{P}$ is a product measure, the basis functions should be mutually orthogonal, and if $\mathbb{P}$ is the uniform measure, they must coincide with the standard parity functions.

\section{Generalized Fourier Representation}\label{sec:generalized}

We come back to the Hilbert space $\mathcal{H}$ of pseudo-Boolean functions equipped with an arbitrary probability measure $\mathbb{P}$. Let $\mathcal X$ be its support. The associated inner product is defined as $\langle f,g \rangle \coloneqq \sum_{\mathbf{x} \in \{0,1\}^d} f(\mathbf{x})g(\mathbf{x})p(\mathbf{x})$, where $p$ denotes the probability mass function (PMF) of $\mathbf{X} \sim \mathbb{P}$. Moreover, the corresponding norm will be denoted by $ \| \cdot \|_{\mathbb P} $. For any subset $S \subseteq [d]$, let $p_S$ denote the marginal PMF of the subvector $\mathbf{X}_S$. In the standard Fourier analysis we have $p_S = 1/2^{\vert S \vert}$ for all $S \subseteq [d]$.

\begin{tcolorbox}[blue_style]
    \begin{definition}[Scaled Parity Functions]
    For any subset $S \subseteq [d]$ and any configuration $\underline{\mathbf{x} \in \mathcal{X}}$, we define the \emph{scaled parity function} $\psi_S$ as follows:
        \begin{equation}
        \psi_S(\mathbf{x}) \coloneqq \frac{ \chi_S(\mathbf{x}) }{2^{\vert S \vert} \cdot  p_S( \mathbf{x}_S ) }.
        \end{equation}
    \end{definition}
\end{tcolorbox}
Here, the term $\chi_S(\mathbf{x})$ is the standard parity function, ensuring the \emph{spanning property}. Then, what we call the \emph{inverse probability weighting} counterbalances the non-uniformity of the measure $\mathbb{P}$ and ensures the hierarchical orthogonality condition \eqref{eq:orthogonality}. Finally, the scaling factor $2^{\vert S \vert}$ acts as a normalization constant and allows to exactly recover the orthonormal Walsh-Hadamard basis in the uniform case.

\subsection{Full Support Setting}
We primarily investigate the case where the distribution $\mathbb{P}$ satisfies the \emph{full support assumption}, \textit{i.e.} when $\mathcal X = \{0,1\}^d$ and for all $\mathbf x \in \{0,1\}^d, p( \mathbf x) > 0$. This obviously ensures that Assumption \ref{ass:weak_dependence} holds. Consequently, we derive the following representation theorem.

\begin{tcolorbox}[blue_style]
\begin{theorem}[Generalized Fourier Representation]\label{thm:gen_fourier}
Under the full support assumption, any pseudo-Boolean function $f \in \mathcal{H}$ admits the unique \emph{generalized Fourier expansion}:
\begin{equation}\label{eq:gen_expansion}
    f(\mathbf{x}) = \sum\limits_{S \subseteq [d]} \widehat{f}(S) \cdot \psi_S(\mathbf{x}).
\end{equation}
\end{theorem}
\end{tcolorbox}
This expansion constitutes the unique explicit solution to the variational problem formulated by \citet{hooker_2007}. Specifically, the terms $f_S(\mathbf{X}_S) \coloneqq \widehat{f}(S) \cdot \psi_S(\mathbf{X})$ depend exclusively on the variables indexed by $S$ and satisfy the hierarchical orthogonality condition \eqref{eq:orthogonality}.

When the measure $\mathbb P$ is product, the collection of function $\left\{ \psi_S \right\}_{S \subseteq [d]}$ becomes mutually orthogonal and the decomposition \eqref{eq:gen_expansion} recovers the standard HFD formulation \eqref{eq:anova_indep}.

\paragraph{Linear Formulation.}
The numbers $\{\widehat{f}(S)\}_{ S \subseteq [d] }$ constitute the \emph{generalized Fourier coefficients} of $f$ with respect to the functions $\psi_S$. They act as the coefficients $\{ \beta_S \}_{ S \subseteq[d] }$ of the following WLS (Weighted Least Squares) regression problem:
\begin{tcolorbox}[blue_style]
\begin{equation}\label{eq:wls}
    \min\limits_{ \bm \beta } \| f - \sum\limits_{S \subseteq [d]} \beta_S \cdot \psi_S \|_{\mathbb P}^2,
\end{equation}
\end{tcolorbox}
\begin{theorem}[Fourier Transform]\label{thm:gen_transform}
    Let $\bm \Psi \in \mathbb{R}^{2^d \times 2^d}$ be the \emph{design matrix} on the Boolean hypercube, with rows indexed by configurations $\mathbf{x} \in \{0,1\}^d$ and columns by subsets $S \subseteq [d]$, defined by:
    \begin{equation}
        \bm \Psi_{\mathbf{x}, S} \coloneqq \psi_S(\mathbf{x}).
    \end{equation}
    The matrix $\bm \Psi$ is invertible. Consequently, the generalized Fourier coefficients are uniquely determined by the linear mapping $\mathcal{F}$ defined as:
    \begin{equation}
        \mathcal{F}(f)(S) \coloneqq \sum\limits_{\mathbf{x} \in \{0,1\}^d} (\bm \Psi^{-1})_{S, \mathbf{x}} \cdot f(\mathbf{x}).
    \end{equation}
    This mapping $\mathcal{F}$ constitutes the Generalized Fourier Transform on $\mathcal{H}$, satisfying $\mathcal{F}(f) = \widehat{f}$.
\end{theorem}

An alternative characterization of the coefficients arises from the geometry of the Hilbert space. By projecting the expansion \eqref{eq:gen_expansion} onto a basis element $\psi_T$ and exploiting the linearity of the inner product, we obtain:
\begin{equation}
    \underbrace{\langle f , \psi_T\rangle}_{ m_T } = \sum\limits_{S \subseteq [d]} \widehat{f}(S) \cdot \underbrace{\langle \psi_S , \psi_T \rangle}_{G_{S,T}}.
\end{equation}
This relation defines a linear system involving the Gram matrix of the basis $\{ \psi_S \}_{ S \subseteq [d] }$ denoted by $G$ and the vector $\bm m$. Since the basis is linearly independent, $G$ is symmetric positive definite and thus invertible. The coefficients are recovered via:
\begin{equation}\label{eq:gram_inversion}
    \widehat{f}(S) = \sum\limits_{T \subseteq [d]} (G^{-1})_{S , T} \cdot m_T.
\end{equation}
The entries of this system are expressed directly as expectations under the \emph{true} distribution $\mathbb{P}$ since we have:
\begin{eqnarray}
    G_{S,T} &=& \mathbb{E}\left[ \psi_S(\mathbf{X}) \psi_T(\mathbf{X}) \right], \\ 
        m_T &=& \mathbb{E}\left[ f(\mathbf{X}) \psi_T( \mathbf{X}) \right].
\end{eqnarray}
Unlike the standard Fourier case where $G$ is the identity matrix, in this general setting, $G$ is a general covariance matrix non necessarily diagonal.

\begin{figure*}[ht] \centering \includegraphics[width=\dimexpr(\sixfigscale\linewidth)/6-1.5pt\relax]{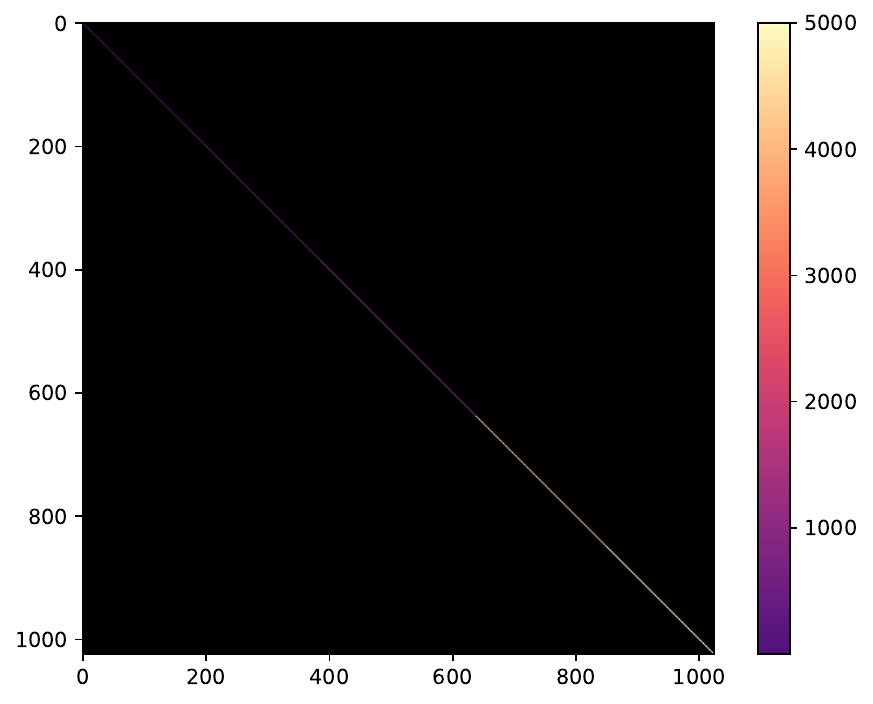}\hfill \includegraphics[width=\dimexpr(\sixfigscale\linewidth)/6-1.5pt\relax]{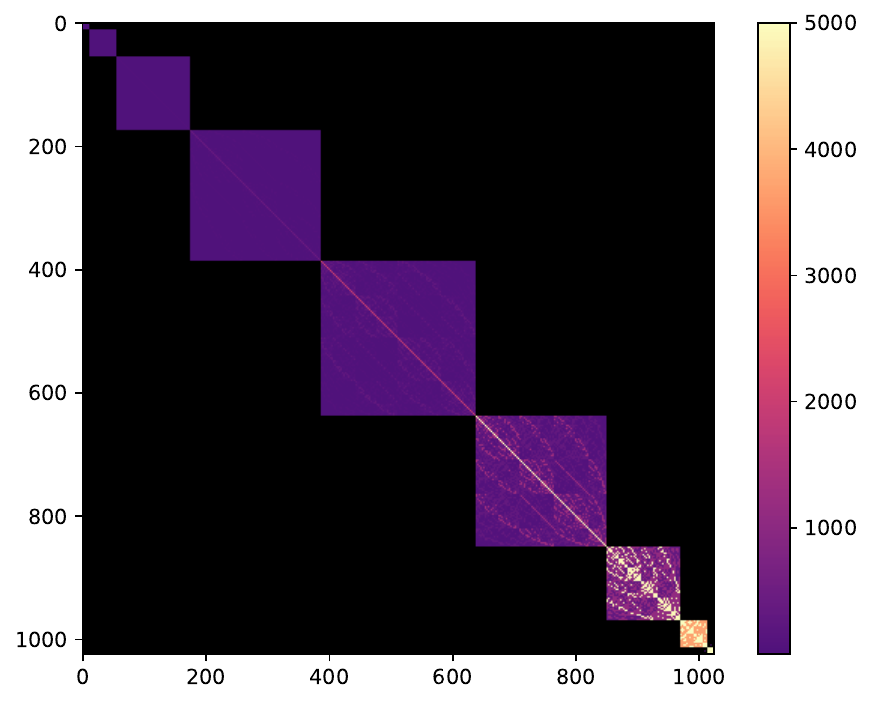}\hfill \includegraphics[width=\dimexpr(\sixfigscale\linewidth)/6-1.5pt\relax]{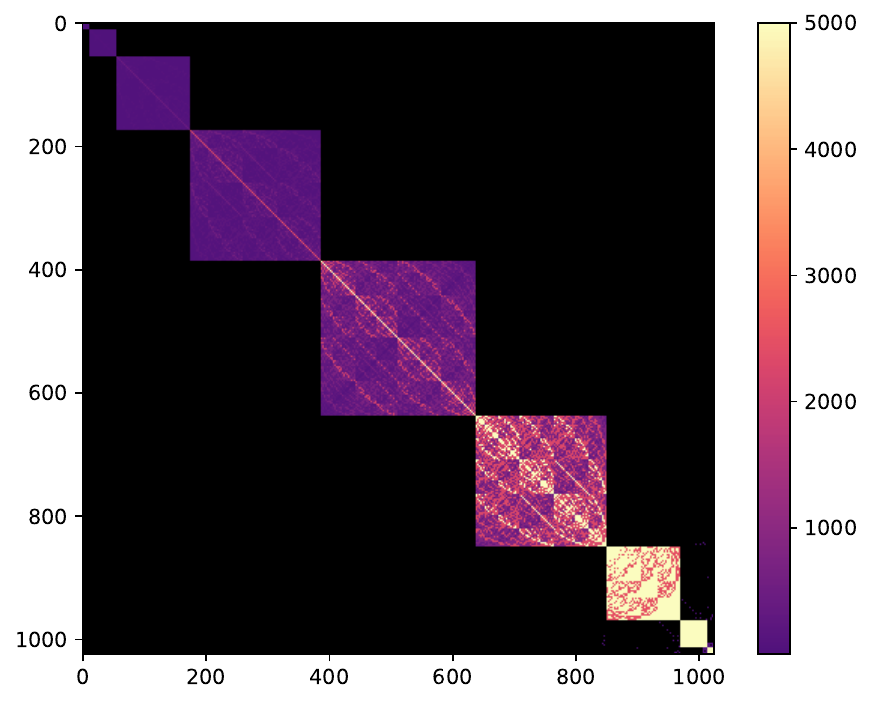}\hfill \includegraphics[width=\dimexpr(\sixfigscale\linewidth)/6-1.5pt\relax]{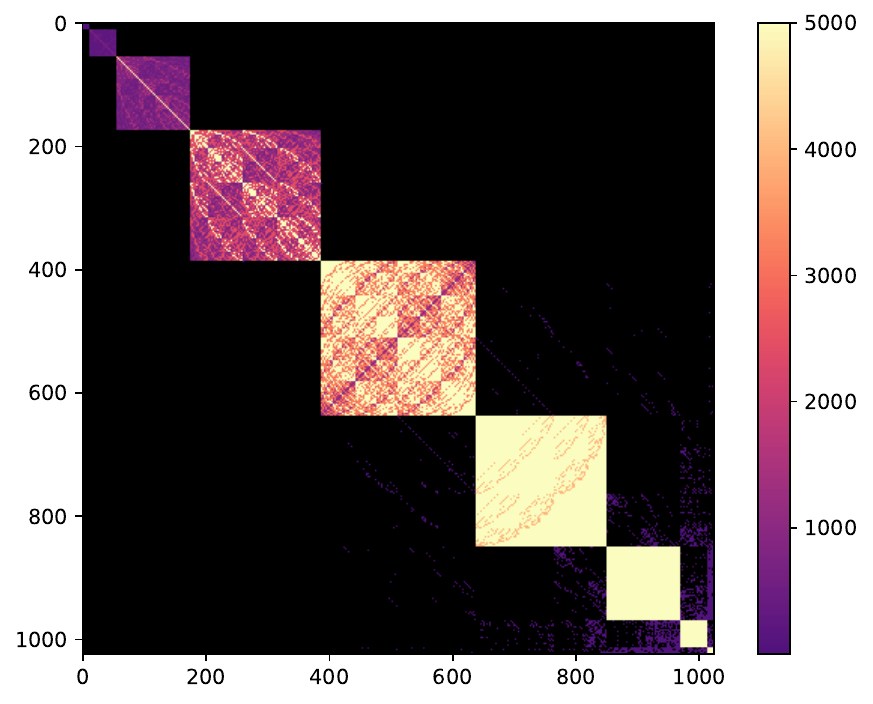}\hfill \includegraphics[width=\dimexpr(\sixfigscale\linewidth)/6-1.5pt\relax]{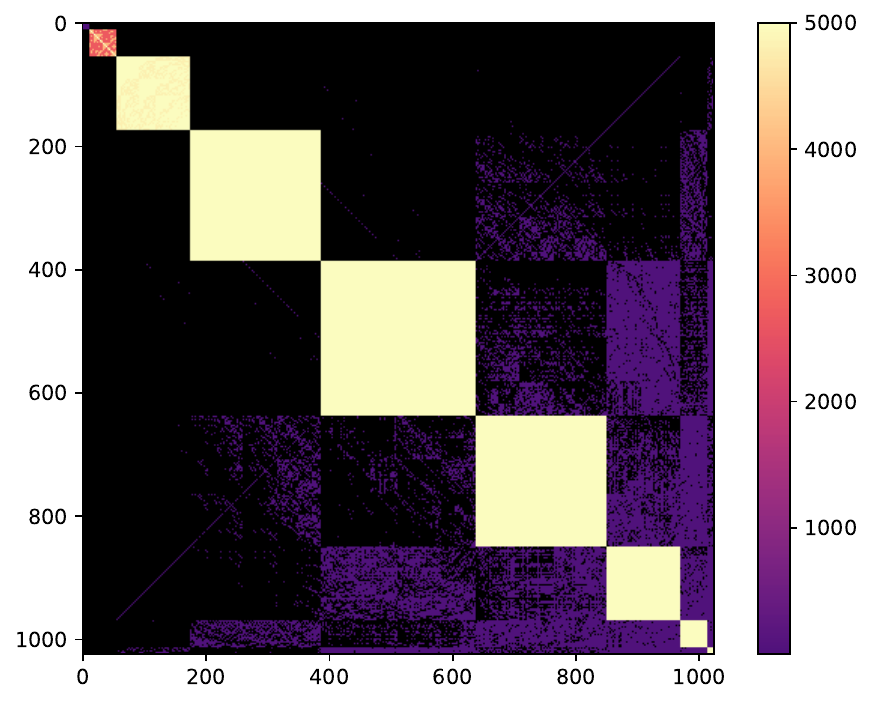}\hfill \includegraphics[width=\dimexpr(\sixfigscale\linewidth)/6-1.5pt\relax]{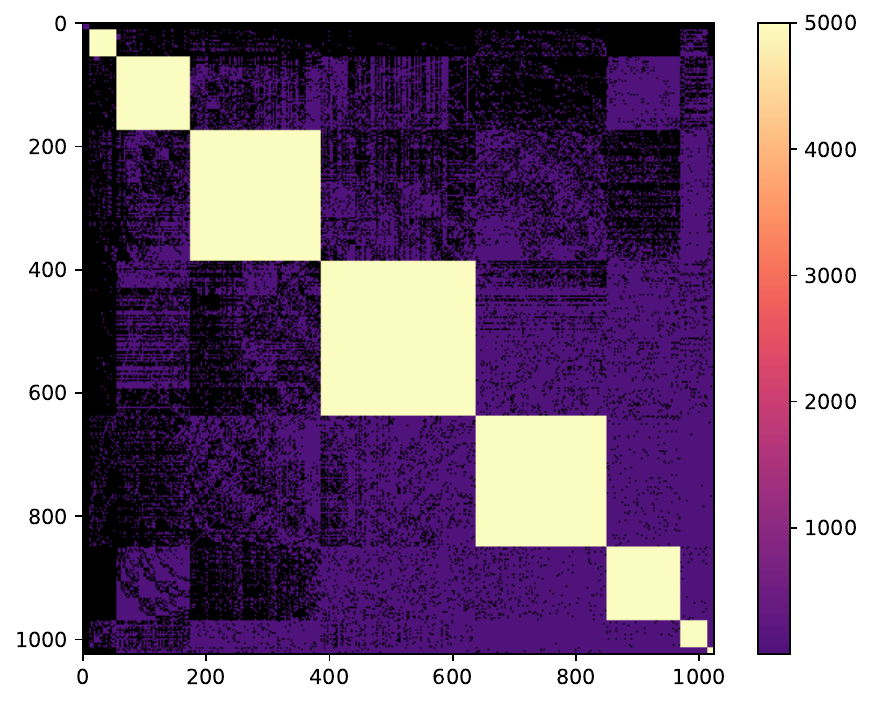} 
\caption{\textbf{Heatmaps of Gram matrices.} Gram matrices $G$ of the generalized Fourier basis for $d=10$ under Bernoulli distribution with increasing a correlation parameter from left to right. At left, $G$ is diagonal, reflecting exact orthogonality of the basis. As the correlation increases, more off-diagonal coefficients appear.}
\label{fig:gamma_plots} 
\end{figure*}

\subsection{Non-Full Support Setting}

While the assumption of full hypercube support guarantees the uniqueness of the HFD, this condition rarely holds in practice. In this section, we address the scenario where the support $\mathcal{X}$ is a strict subset of the hypercube. This case is very common in standard machine learning, especially with binary variables when dealing with one-hot encoded features or small number of samples in the dataset. In these cases, the corresponding measure $\mathbb P$ is sparse by construction, leaving the vast majority of the hypercube unobserved regardless of the population distribution.

\begin{tcolorbox}[blue_style]
\paragraph{Key Message.}
We must acknowledge that, in this regime, the existing literature remains both incomplete and somewhat restrictive. Early foundational works—most notably \citet{stone1994use} and \citet{hooker_2007}—define the HFD under a hyperrectangular support together with a bounded probability density. \citet{chastaing_generalized_2012} subsequently relaxed this requirement via Assumption~\ref{ass:weak_dependence}. While these conditions align naturally with the continuous setting, they are typically violated (or simply inapplicable) in discrete scenarios.
In what follows, we therefore propose a strategy based on the regularization of \eqref{eq:wls} to recover \emph{a} decomposition in the non-full support setting. This case is arguably the most prevalent in practice, yet it is also among the least systematically treated in the literature. Concretely, we introduce a heuristic that preserves the ANOVA formulation~\eqref{eq:def_anova} as well as the hierarchical orthogonality~\eqref{eq:orthogonality}, by expressing the components through the functions $\psi_S$. To date, however, there is no canonical, unified extension that would single out a uniquely \emph{best} decomposition in this general setting.
\end{tcolorbox}

\paragraph{Regularized Weighted Least Square Problem.}
To address this ambiguity and recover a meaningful representation, we must impose structural priors on the solution, typically sparsity. This requires framing the decomposition as a penalized optimization problem. Specifically, we employ regularized WLS regression, encompassing methods such as LASSO, Ridge, or Elastic Net.

Formally, let us define the reconstruction error with respect to the coefficient vector $\bm{\beta}$ as:
\begin{equation}
    \mathcal{L}(\bm{\beta}) \coloneqq \| f - \sum_{S \subseteq [d]} \beta_S \cdot \psi_S \|_{\mathbb P}^2,
    \label{eq:wls_loss}
\end{equation}
which corresponds to the objective function of the standard WLS problem \eqref{eq:wls}. To enforce uniqueness and sparsity, we define the following penalized minimization problem:
\begin{equation}
    \min_{\bm{\beta}} \left\{ \mathcal{L}(\bm{\beta}) + \mathrm{pen}( \bm \beta) \right\},
    \label{eq:penalized_wls}
\end{equation}
where $\mathrm{pen}(\bm{\beta})$ is the regularization term. In practice, a natural default for penalized WLS is the Elastic Net, as it combines the sparsity-inducing effect of the $\ell_1$ penalty with the stability and grouping behavior of an $\ell_2$ penalty in the presence of correlated (and often redundant) features. This is particularly relevant in expansions where many $\psi_S$ can be collinear, for which pure LASSO can yield unstable model selection, while Ridge remains dense. Moreover, adding a strictly positive $\ell_2$ component makes the objective strongly convex, ensuring a unique minimizer and improved numerical conditioning.

\subsection{Empirical Perspective}
In this subsection, we formalize the empirical framework and discuss the computational challenges inherent to the resulting minimization problem.

\paragraph{Empirical Setting.}We consider a dataset $\mathcal{D}$ consisting of $n$ \emph{distinct} observations $\{ \mathbf{x}^{(1)}, \dots, \mathbf{x}^{(n)} \} \subseteq \{0, 1\}^d$ from the Boolean hypercube, which is almost always the case in boolean or categorical standard setting. Since the observations are unique, $\mathcal{D}$ constitutes the exact support of the empirical distribution. Consequently, the empirical probability mass function assigns a uniform mass to each sample:
\begin{equation} 
\widehat{p}_n(\mathbf{x}) = \frac{1}{n} \cdot \mathbf{1}_{\mathcal{D}}(\mathbf{x}),
\end{equation}
where $\mathbf{1}_{\mathcal{D}}(\cdot)$ denotes the indicator function of the set $\mathcal{D}$. In high-dimensional settings, it is typical that $n \ll 2^d$. This implies that the probability mass is concentrated on a negligible fraction of the hypercube. As a result, the effective dimension of the space of pseudo-Boolean functions under this empirical measure is significantly lower than the ambient dimension $2^d$.

\newcommand{\gridscale}{0.8}
\begin{figure*}[t!] 
    \centering

    \begin{subfigure}[b]{\gridscale\dimexpr0.32\textwidth\relax}
        \centering
        \includegraphics[width=\linewidth]{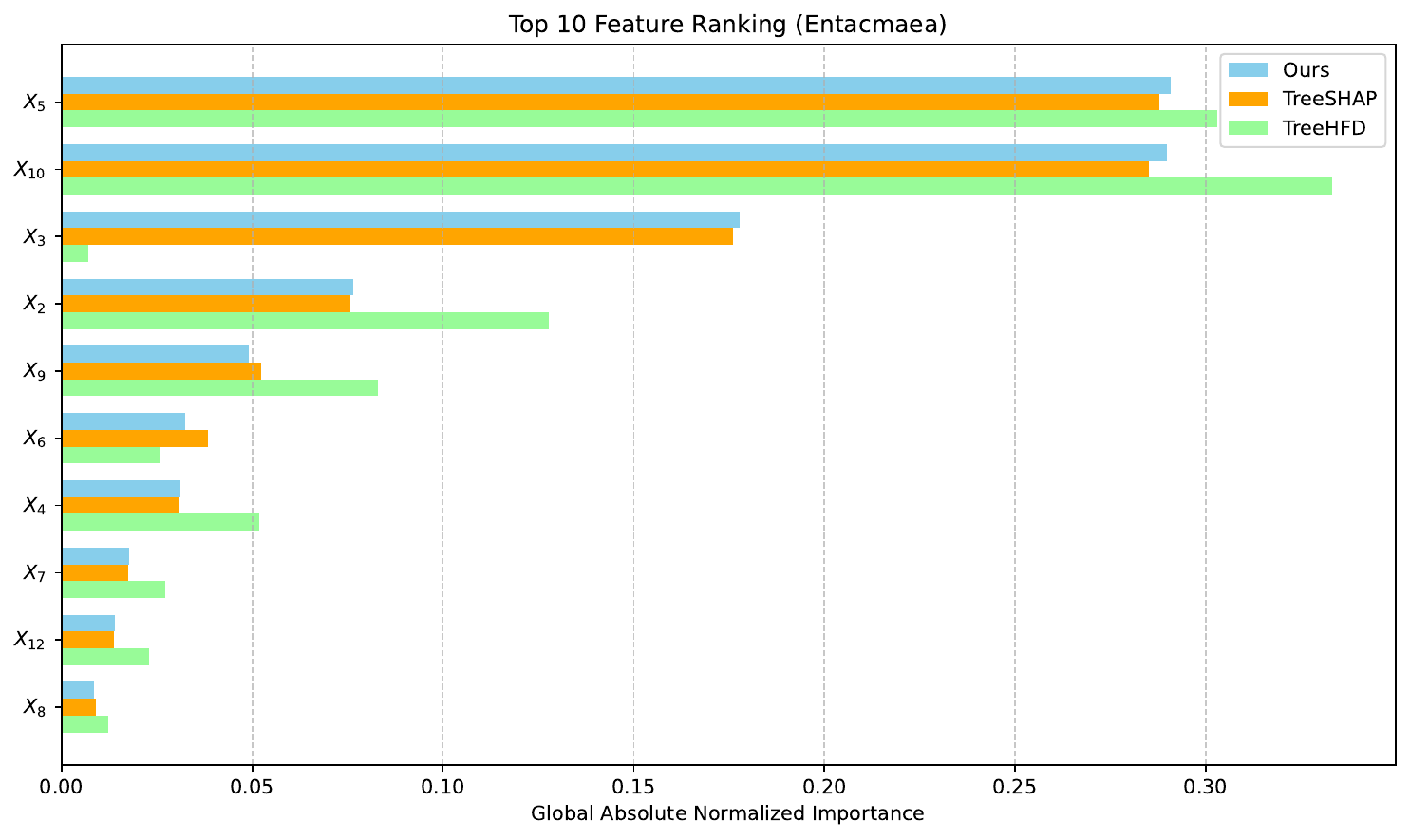}
        \caption{Entacmaea (A)}
        \label{subfig:entacmaea}
    \end{subfigure}
    \hfill
    \begin{subfigure}[b]{\gridscale\dimexpr0.32\textwidth\relax}
        \centering
        \includegraphics[width=\linewidth]{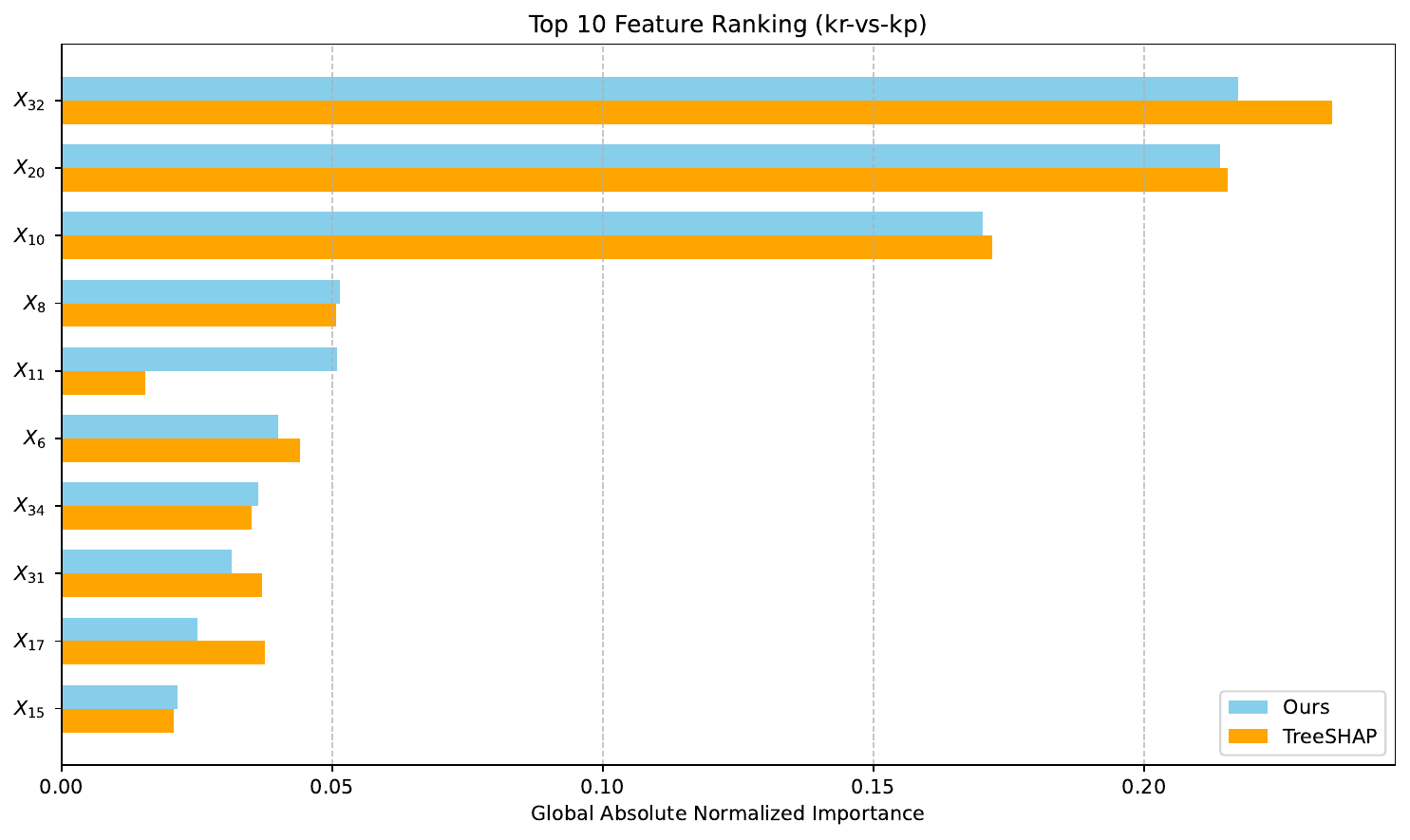}
        \caption{kr-vs-kp (B)}
        \label{subfig:krvskp}
    \end{subfigure}
    \hfill
    \begin{subfigure}[b]{\gridscale\dimexpr0.32\textwidth\relax}
        \centering
        \includegraphics[width=\linewidth]{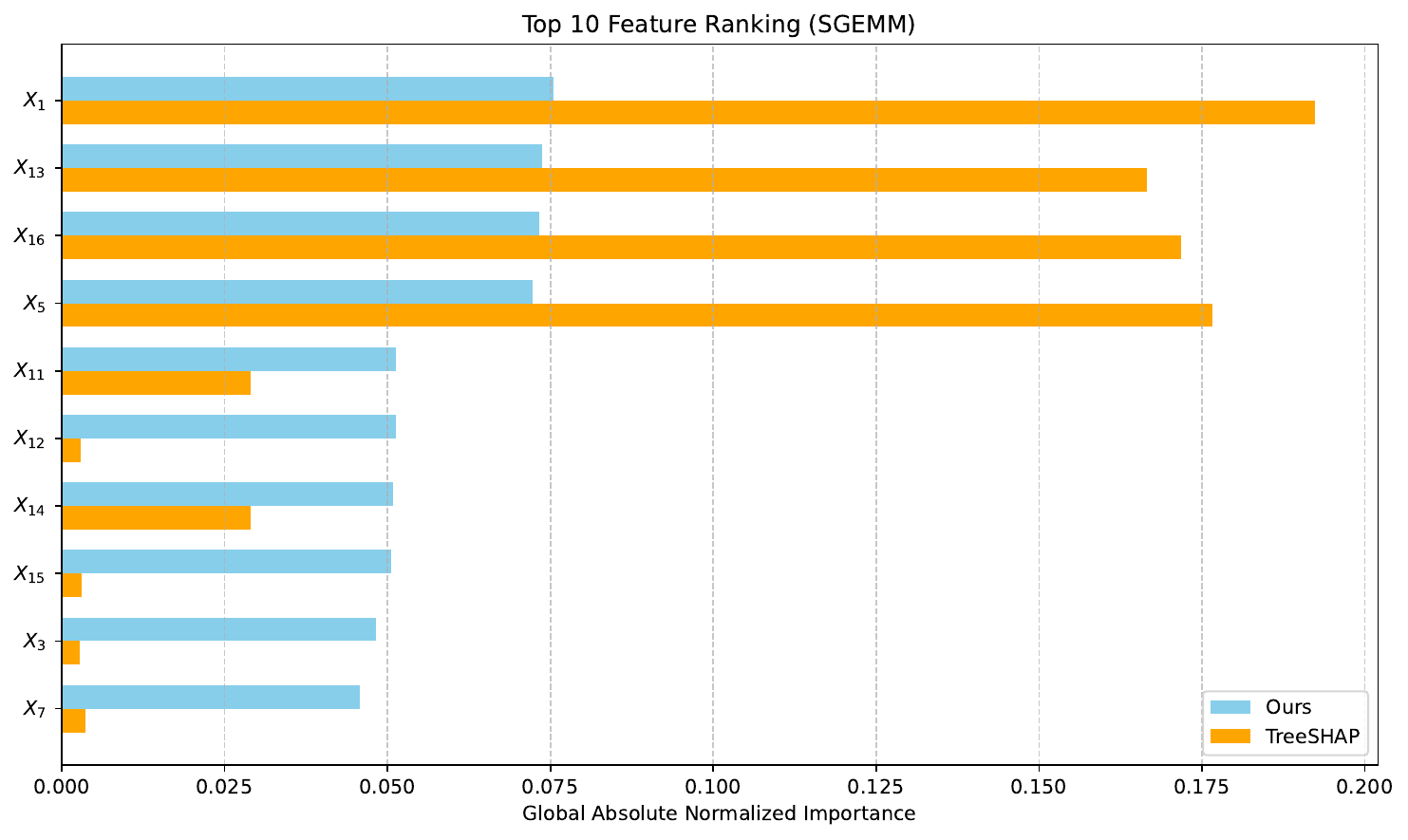}
        \caption{SGEMM (C)}
        \label{subfig:sgemm}
    \end{subfigure}

    \vspace{0.5cm} 

    \begin{subfigure}[b]{\gridscale\dimexpr0.32\textwidth\relax}
        \centering
        \includegraphics[width=\linewidth]{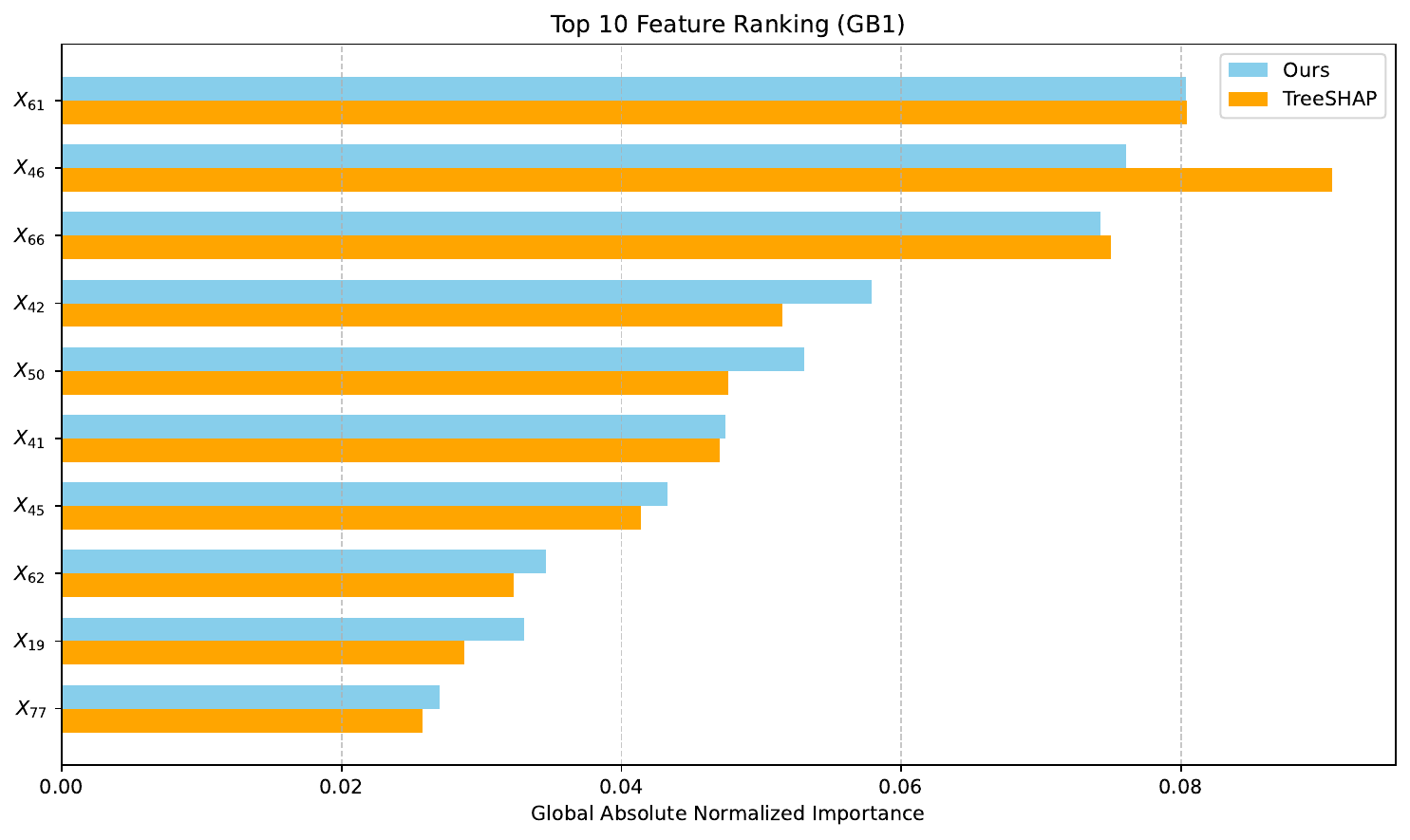}
        \caption{GB1 (D)}
        \label{subfig:gb1}
    \end{subfigure}
    \hfill
    \begin{subfigure}[b]{\gridscale\dimexpr0.32\textwidth\relax}
        \centering
        \includegraphics[width=\linewidth]{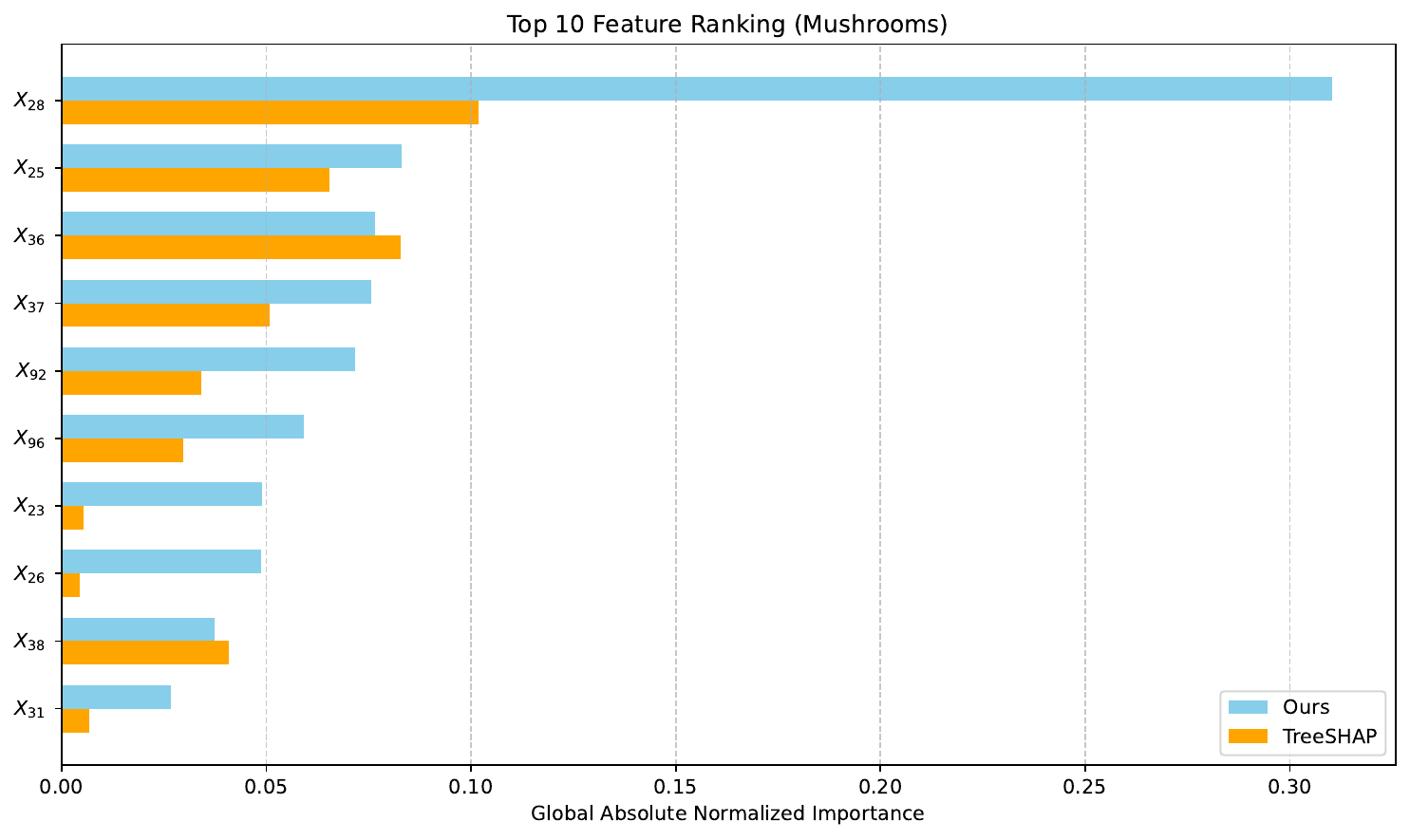}
        \caption{Mushrooms (E)}
        \label{subfig:mushrooms}
    \end{subfigure}
    \hfill
    \begin{subfigure}[b]{\gridscale\dimexpr0.32\textwidth\relax}
        \centering
        \includegraphics[width=\linewidth]{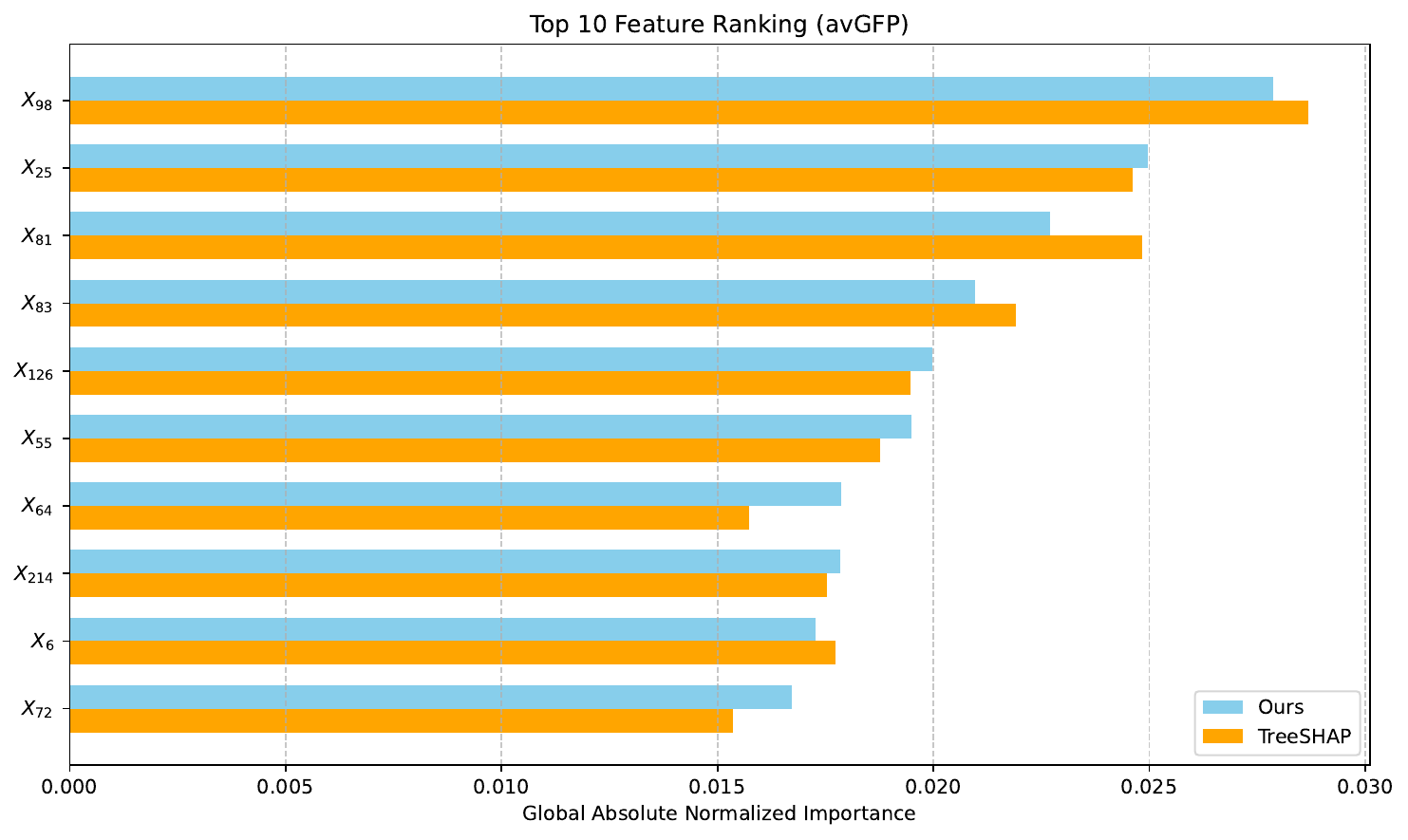}
        \caption{avGFP (F)}
        \label{subfig:avgfp}
    \end{subfigure}

    \caption{\textbf{Global feature importance on tree-based models.} For each dataset, we report the top-10 features ranked by global importance according to our method, comparing TreeSHAP (and TreeHFD on Entacmaea only) to our method. Across all six datasets, the induced attributions are consistent: the rankings and values are very close.}
    \label{fig:tree}
\end{figure*}

\paragraph{Empirical Risk Minimization.}
Under the empirical measure $\widehat{p}_n$, the objective function reduces to the standard Ordinary Least Squares (OLS) loss:
\begin{equation}
    \frac{1}{n} \underbrace{\sum\limits_{ i=1 }^{n} \left( f( \mathbf{x}^{(i)} ) - \sum_{S \subseteq [d]} \beta_S \cdot \psi_S( \mathbf{x}^{(i)} ) \right)^2}_{ \| {\bm y}_n - \bm \Psi \bm \beta \|^2_{ \mathbb{R}^n } },
\end{equation}
where ${\bm y}_n = [f(\mathbf{x}^{(1)}), \dots, f(\mathbf{x}^{(n)})]^\top \in \mathbb{R}^n$ is the target vector, and $\mathbf{\Psi} \in \mathbb{R}^{n \times 2^d}$ is the design matrix with entries $\psi_S(\mathbf{x}^{(i)})$. Although the sample size $n$ may be small, this optimization remains computationally intractable in its primal form, as it involves a design matrix $\mathbf{\Psi}$ with an exponential number of columns.

\paragraph{Low-Order Approximation.}
To address the computational intractability of the full basis expansion, we restrict the model to a subspace of bounded complexity. Specifically, we employ a \emph{truncated} expansion where coefficients $\beta_S$ are constrained to zero for all interactions $|S| > k$. While the full design matrix scales exponentially with dimension $d$, this truncation reduces the complexity to $\mathcal{O}(d^k)$, rendering the problem tractable. In tabular domains, it is standard to set $k=2$, leveraging the \emph{sparsity of effects} principle \citep{montgomery2017design} which assumes that signal energy is concentrated in main effects and pairwise interactions \citep{yu2019reluctant,lengerich2020purifying,benard2025tree}. More generally, one can select a subset of $r \ll 2^d$ columns comprising all low-order terms—to form a \emph{reduced design matrix} $\mathbf{\Psi}_{red} \in \mathbb{R}^{n \times r}$. Consequently, we formulate the estimator as the following penalized least squares optimization:
\begin{tcolorbox}[blue_style]
\begin{equation}\label{eq_empirical}
\min\limits_{ \bm \beta \in \mathbb{R}^{r} } \frac{1}{n} \| \bm y_n - \bm \Psi_{red} \bm \beta \|_{ \mathbb{R}^n }^2 + \mathrm{pen}( \bm \beta ).
\end{equation}
\end{tcolorbox}

\section{Numerical Experiments}\label{sec:numeric}

\paragraph{Experimental Setup.}
We evaluate our framework on six real-world datasets, summarized in Table~\ref{tab:datasets}. To ensure consistency, all categorical features were one-hot encoded to yield strictly boolean inputs, and constant features were pruned post-binarization.
For each dataset, we define the pseudo-Boolean function $f$ by training a standard machine learning model in a \emph{black-box} context. We employed Random Forests for classification tasks (\textbf{B}, \textbf{E}). For regression tasks, we trained both eXtreme Gradient Boosting (XGB) and Multi-Layer Perceptrons (MLP). Training details, architectures, and hyperparameters are provided in the Appendix.

\begin{table}[ht]
\caption{ Datasets. The task is either Classification (C) or Regression (R). $d$ denotes the feature dimension after binarization.}
\label{tab:datasets}
\centering
\resizebox{\columnwidth}{!}{
\begin{tabular}{llcrr}
\toprule
ID & Dataset & Task & $d$ & $n$\\
\midrule
\textbf{A} & Entacmaea \citep{poelwijk2019learning} & R & 13 & 8\,192 \\
\textbf{B} & kr-vs-kp \citep{dua2017uci} & C & 35 & 3\,196 \\
\textbf{C} & SGEMM \citep{dua2017uci} & R & 40 & 241\,600 \\
\textbf{D} & GB1 \citep{wu2016adaptation} & R & 80 & 149\,361 \\
\textbf{E} & Mushrooms \citep{dua2017uci} & C & 116 & 8\,124 \\
\textbf{F} & avGFP \citep{sarkisyan2016local} & R & 233 & 54\,025 \\
\bottomrule
\end{tabular}
}
\end{table}

\paragraph{Penalization.}
We solve the empirical problem \eqref{eq_empirical} using Elastic Net regularization. The penalty term is defined as:
\begin{equation}
    \mathrm{pen}( \bm \beta) \coloneqq \alpha \left( \lambda \| \bm \beta \|_1 + \frac{1-\lambda}{2} \| \bm \beta \|_2^2 \right),
\end{equation}
where $\alpha > 0$ controls the regularization intensity and $\lambda \in [0, 1]$ is the mixing parameter between LASSO ($\ell_1$) and Ridge ($\ell_2$) penalties. The hyperparameters taken in our experiments are detailed in the Appendix. 

\paragraph{Approximation Fidelity.}
We assess the quality of the low-order approximation by measuring how well the truncated expansion matches the \emph{black-box} predictor $f$ on the empirical data distribution. Concretely, we compute the reconstruction fidelity
$R^2_{\text{Fourier}}$. Importantly, $R^2_{\text{Fourier}}$ is computed \emph{globally over the full dataset} and quantifies fidelity to $f$ (not predictive performance w.r.t. ground-truth targets). For classification tasks, we compute $R^2_{\text{Fourier}}$ on the predicted probabilities of 
Class 1.

\begin{table}[ht]
\caption{Table of performances.}
\label{tab:scores}
\centering
\begin{tabular}{l c cc cc cc}
\toprule
 &  & \multicolumn{2}{c}{Model Score} & \multicolumn{2}{c}{$R^2_{\text{Fourier}}$} & \multicolumn{2}{c}{Time (s)} \\
\cmidrule(lr){3-4} \cmidrule(lr){5-6} \cmidrule(lr){7-8}
ID & $k$ & Tree & MLP & Tree & MLP & Tree & MLP \\
\midrule
\textbf{A} & 2 & 0.96 & 0.95 & 0.93 & 0.93 & 0.03 & 0.05 \\
\textbf{B} & 1 & 0.93 & --   & 0.90 & --   & 0.02 & -- \\
\textbf{C} & 2 & 0.94 & 0.99 & 0.87 & 0.79 & 35 & 37 \\
\textbf{D} & 2 & 0.86 & 0.96 & 0.66 & 0.66 & 118 & 122 \\
\textbf{E} & 1 & 0.99 & --   & 1.00 & --   & 0.18 & -- \\
\textbf{F} & 1 & 0.59 & 0.72 & 0.97 & 0.72 & 0.65 & 0.65 \\
\bottomrule
\end{tabular}
\end{table}
The metric $R^2_{\text{Fourier}}$ quantifies the \emph{fidelity} of the expansion with respect to the black-box $f$: a value of $R^2_{\text{Fourier}} \approx 1$ implies that the truncated expansion perfectly reproduces the function $f$. The metric Time report the time in seconds to obtain the decomposition (see Appendix for more details).

\newcommand{\mlpscale}{0.8} 

\begin{figure*}[t!]
  \centering

  \begin{subfigure}[b]{\mlpscale\dimexpr0.49\textwidth\relax}
    \centering
    \includegraphics[width=\linewidth]{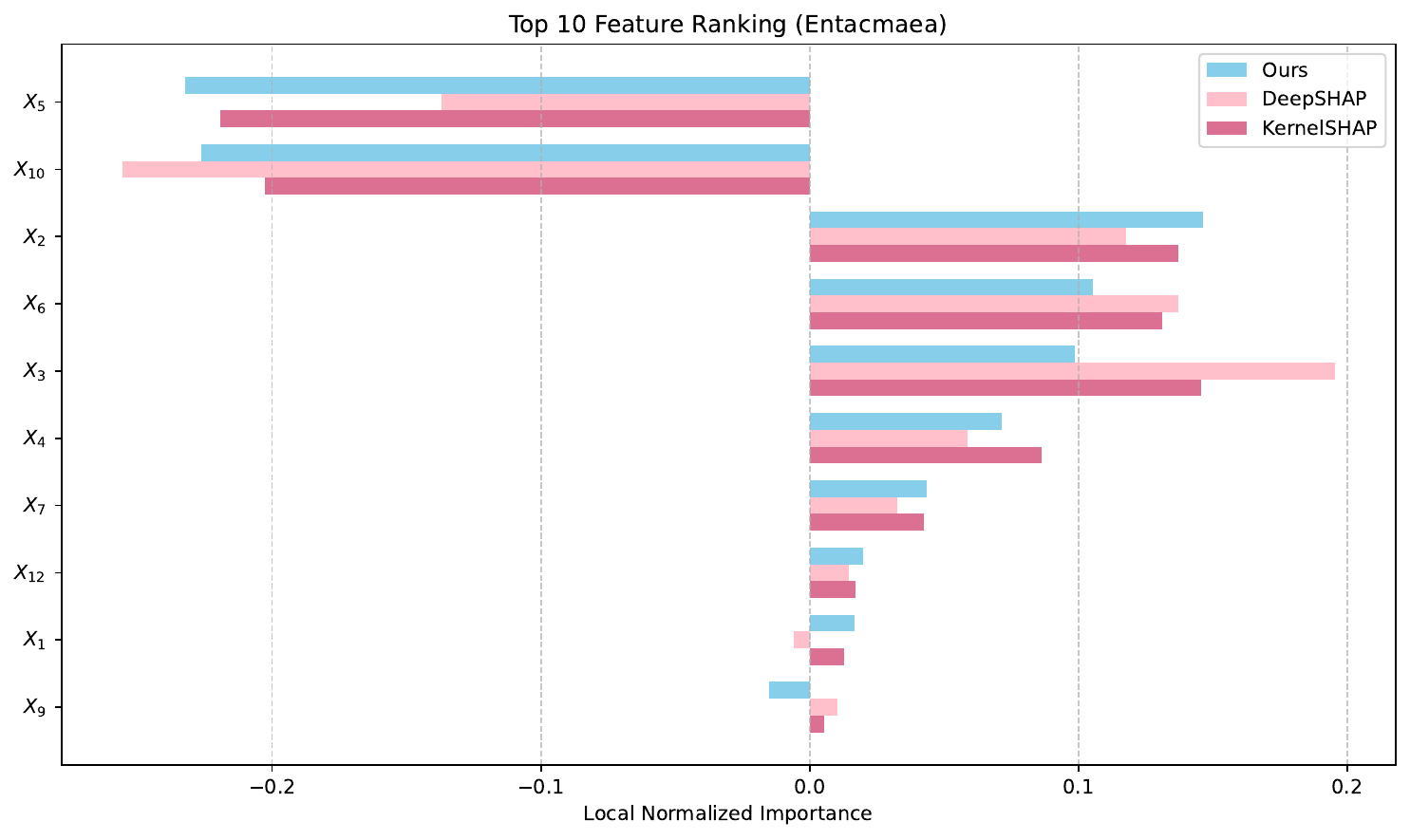}
    \caption{Entacmaea (A)}
    \label{subfig:entacmaea_mlp}
  \end{subfigure}
  \begin{subfigure}[b]{\mlpscale\dimexpr0.49\textwidth\relax}
    \centering
    \includegraphics[width=\linewidth]{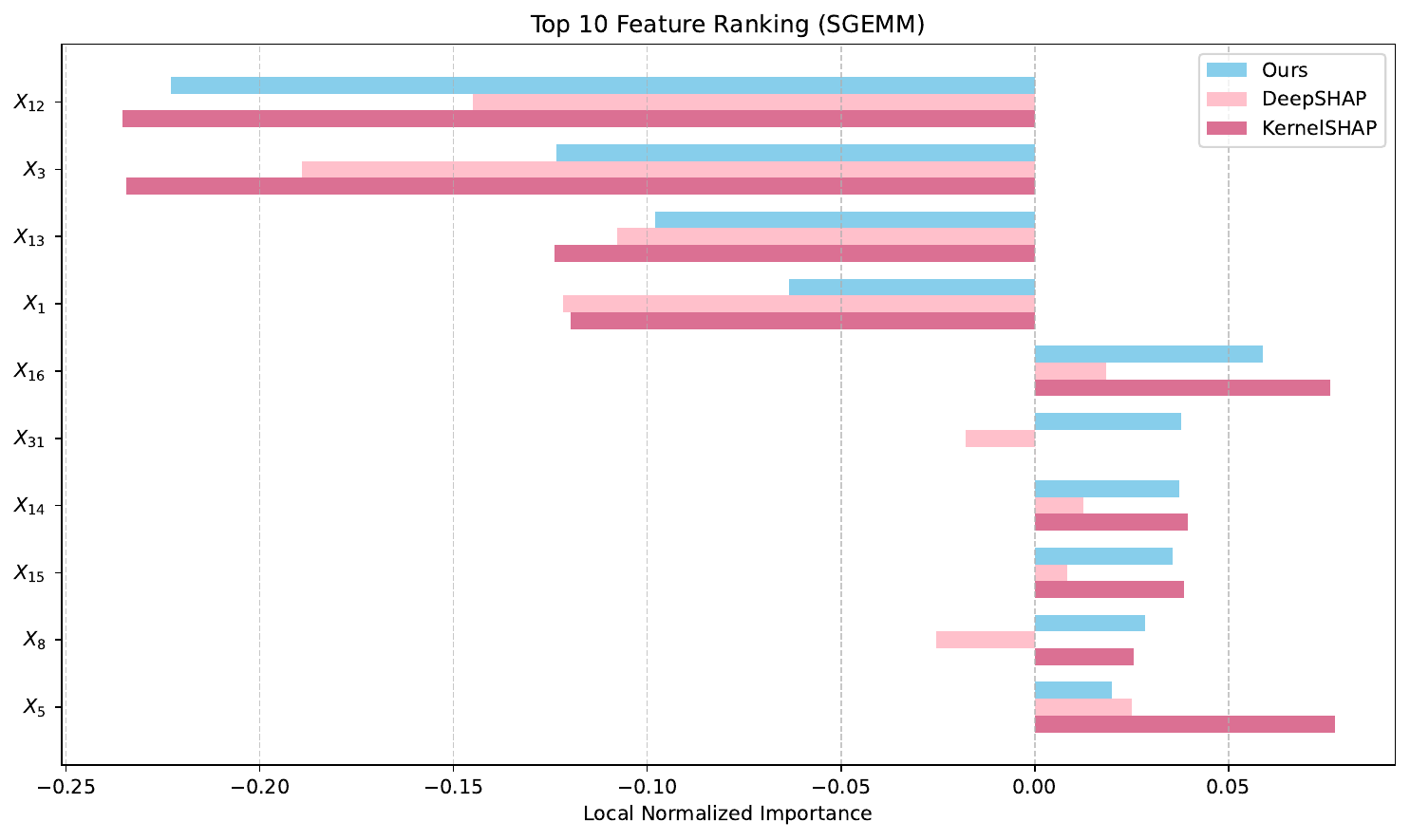}
    \caption{SGEMM (C)}
    \label{subfig:sgemm_mlp}
  \end{subfigure}

  \vspace{0.30cm}

  \begin{subfigure}[b]{\mlpscale\dimexpr0.49\textwidth\relax}
    \centering
    \includegraphics[width=\linewidth]{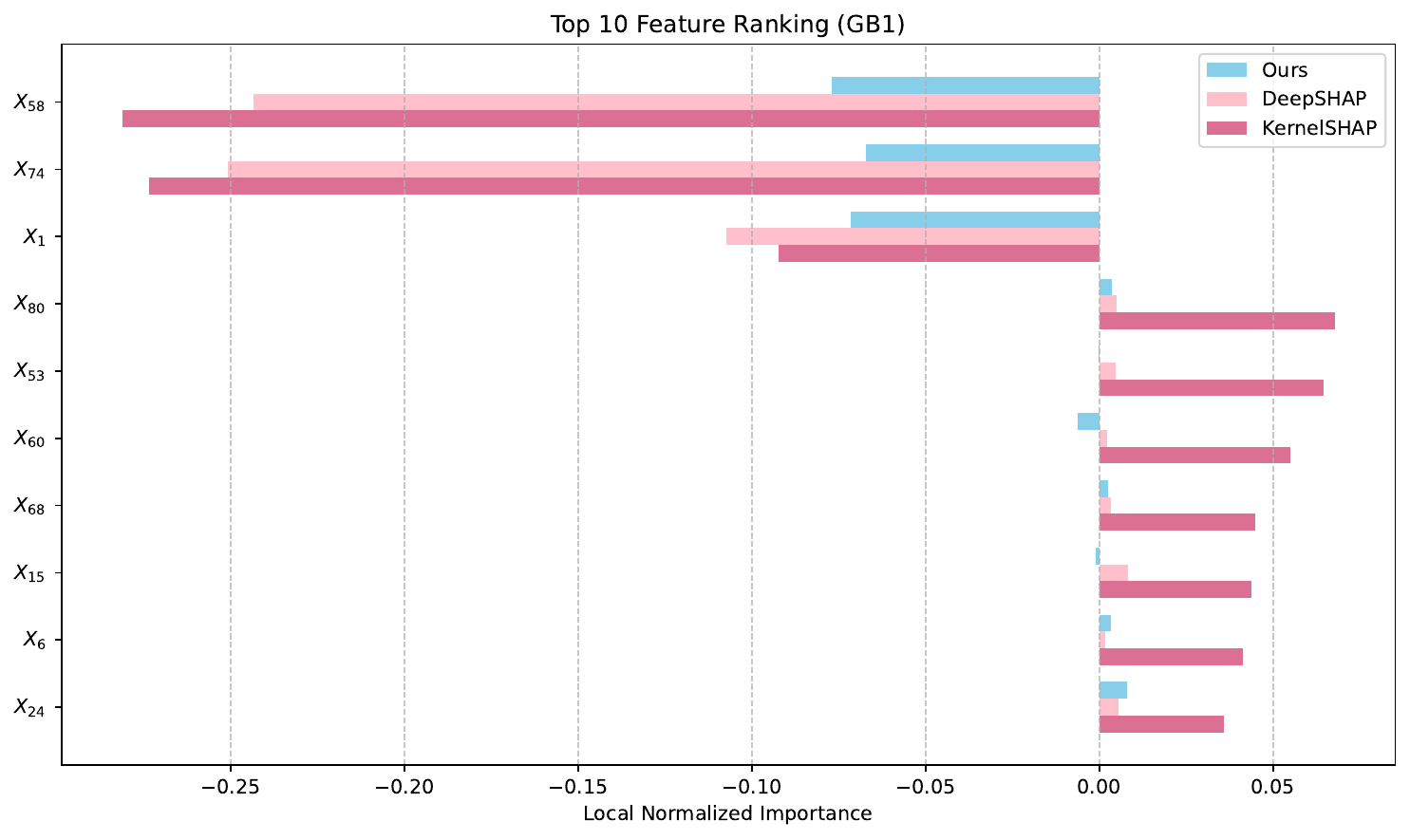}
    \caption{GB1 (D)}
    \label{subfig:gb1_mlp}
  \end{subfigure}
  \begin{subfigure}[b]{\mlpscale\dimexpr0.49\textwidth\relax}
    \centering
    \includegraphics[width=\linewidth]{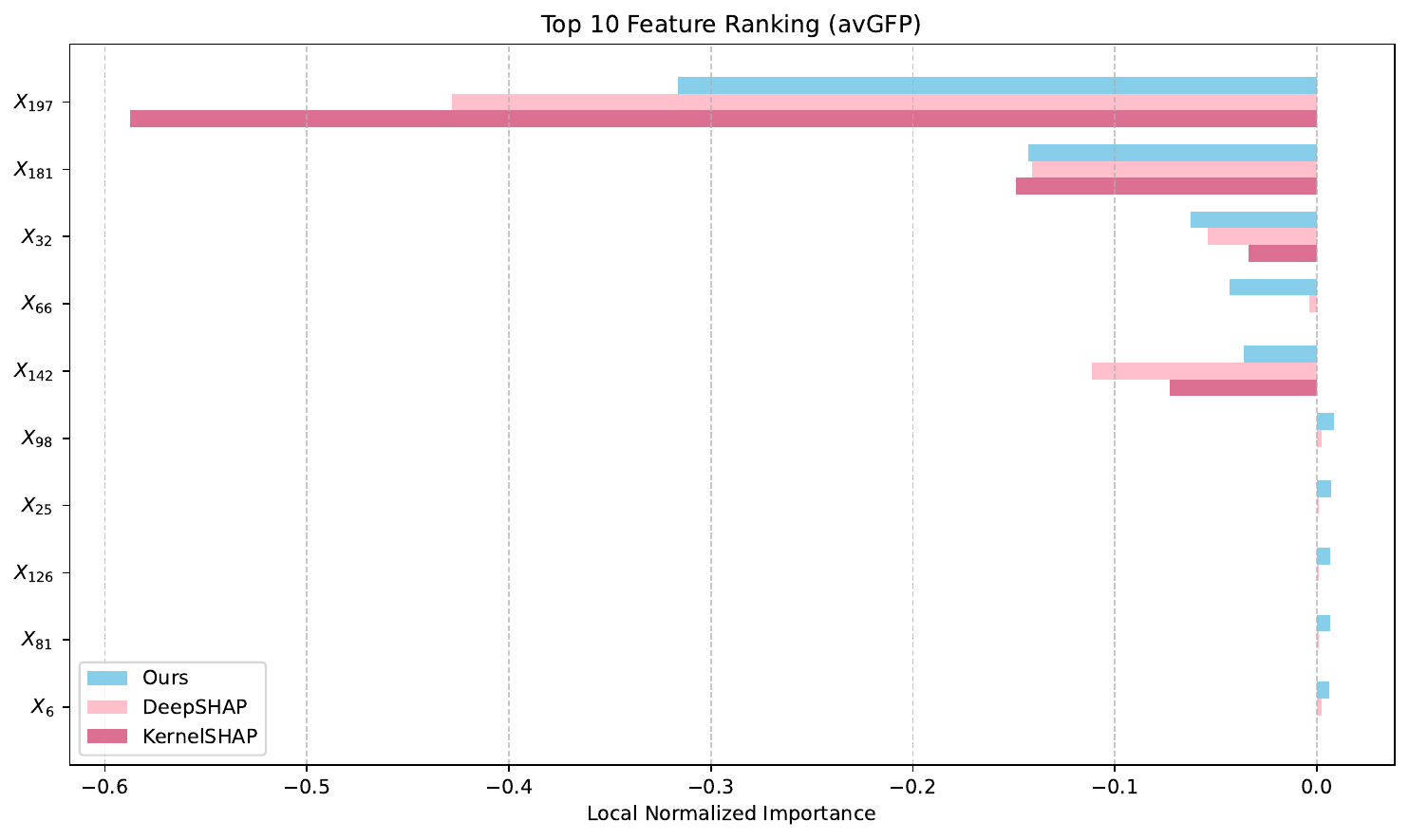}
    \caption{avGFP (F)}
    \label{subfig:avgfp_mlp}
  \end{subfigure}

  \caption{\textbf{Local feature importance on MLP models.}
For each regression task dataset, we report the top-10 locally most influential features for an MLP (based on our method), comparing to DeepSHAP and KernelSHAP on the same randomly selected test point. Bars show signed, locally normalized contributions (positive values increase the prediction, negative values decrease it).}
  \label{fig:mlp}
\end{figure*}

\paragraph{Reconstruction Error.}
Table~\ref{tab:scores} reports both the predictive performance of the black-box model $f$ (\emph{Model Score}) and the fidelity of our truncated Fourier expansion to $f$ ($R^2_{\text{Fourier}}$). Across most datasets, a low-order expansion (with $k\in\{1,2\}$) achieves high fidelity: first- and second-order terms already capture a great part of the variance of $f$, which is consistent with the \emph{sparse effects} hypothesis where higher-order interactions contribute marginally to the decision rule.

Several cases are particularly interesting. On \textbf{E}, a purely additive approximation ($k=1$) yields essentially perfect reconstruction ($R^2_{\text{Fourier}}=1.00$), aligning with prior knowledge that the signal is dominated by a single feature that our framework highly recovers. On \textbf{F}, the model is reconstructed with very high fidelity ($R^2_{\text{Fourier}}=0.97$), this particularly aligns with the low accuracy of the model : this model probably captured only low order additive effects, that we perfectly recovered, but it is doesn't suffice to decompose the entire underlying mapping.

\paragraph{Feature Attribution.}
We compare our attributions to standard SHAP-based baselines across two model families. For tree-based models, we report global importance rankings and compare to TreeSHAP~\cite{lundberg2018consistent} (and TreeHFD~\cite{benard2025tree} when applicable) (Fig.~\ref{fig:tree}). For MLPs, we illustrate local explanations on a randomly selected test instance and compare to KernelSHAP and DeepSHAP~\cite{lundberg2017unified} (Fig.~\ref{fig:mlp}).  Beyond qualitative agreement, our results empirically suggest that, in many practical settings, the feature attributions produced by SHAP are closely connected to a HFD-like decomposition. In the independent setting, this connection is well established and can be made explicit in our framework. Interestingly, we observe that a similar alignment persists on several real-world datasets exhibiting substantial feature dependence, indicating that SHAP attributions may still behave as proxies for measure-dependent low-order effects captured by our truncated expansion. Full computational details are provided in the Appendix. 

\section{Discussion}\label{sec:discuss}

\paragraph{Conclusion.}
We established a formal connection between Fourier analysis on the Boolean hypercube and the Hoeffding (ANOVA) Functional Decomposition (HFD), showing that the former arises as a special case of the latter under a uniform product measure. Building on this insight, we proposed a principled generalization of Fourier-based analysis grounded in the HFD framework, and derived an explicit functional basis for the decomposition that recovers the formulation of \citet{hooker_2007}. A key contribution of this work is the reformulation of the decomposition problem as a weighted least-squares regression, which reduces a nonparametric estimation task to a linear problem. To mitigate the curse of dimensionality, we introduced a regularization scheme that yields accurate low-rank approximations, as confirmed by our empirical evaluation. Finally, we drew connections between our framework and established explainability methods, including SHAP and TreeHFD, highlighting structural similarities in the feature attribution task and suggesting promising directions for future work at the interface of functional decomposition and interpretable machine learning.

\paragraph{Limitations \& Future Work.}
Our framework currently focuses on pseudo-Boolean functions over discrete domains; extending it to continuous features is a primary direction for future work. A more fundamental open problem concerns the non-full-support setting, where many decompositions are consistent with the same high-level desiderata, each reflecting implicit modeling choices. Identifying a canonical selection principle—grounded in invariance, stability, or statistical optimality—remains an important open question. More broadly, we believe that HFD represents a promising and underexplored avenue for both explainability and learning tasks: by making the decomposition basis explicit, a wide range of such tasks reduce to tractable linear problems, opening the door to principled, scalable methods that the ML community has yet to fully exploit.

\newpage
\begin{acknowledgements}
This work was supported by the French Association Nationale de la Recherche et de la Technologie (ANRT) through a CIFRE PhD project at \'Electricité de France (EDF). Fabrice Gamboa and Jean-Michel Loubes are supported by the ANR-3IA Artificial and Natural Intelligence Toulouse Institute (ANITI).
\end{acknowledgements}

\bibliography{biblio}

\begin{thebibliography}{35}
\providecommand{\natexlab}[1]{#1}
\providecommand{\url}[1]{\texttt{#1}}
\expandafter\ifx\csname urlstyle\endcsname\relax
  \providecommand{\doi}[1]{doi: #1}\else
  \providecommand{\doi}{doi: \begingroup \urlstyle{rm}\Url}\fi

\bibitem[B{\'e}nard(2025)]{benard2025tree}
Cl{\'e}ment B{\'e}nard.
\newblock Tree ensemble explainability through the hoeffding functional decomposition and treehfd algorithm.
\newblock \emph{Advances in Neural Information Processing Systems}, 2025.

\bibitem[Bordt and von Luxburg(2023)]{bordt2023shapley}
Sebastian Bordt and Ulrike von Luxburg.
\newblock From shapley values to generalized additive models and back.
\newblock In \emph{International Conference on Artificial Intelligence and Statistics}, pages 709--745. PMLR, 2023.

\bibitem[Chastaing et~al.(2012)Chastaing, Gamboa, and Prieur]{chastaing_generalized_2012}
Gaëlle Chastaing, Fabrice Gamboa, and Clémentine Prieur.
\newblock {Generalized {H}oeffding-Sobol Decomposition for Dependent Variables – Application to Sensitivity Analysis}.
\newblock \emph{Electronic Journal of Statistics}, 6:\penalty0 2420--2448, March 2012.
\newblock \doi{10.1214/12-EJS749}.
\newblock URL \url{https://projecteuclid.org/euclid.ejs/1356098617}.

\bibitem[Covert and Lee(2021)]{covert2021improving}
Ian Covert and Su-In Lee.
\newblock Improving kernelshap: Practical shapley value estimation using linear regression.
\newblock In \emph{International conference on artificial intelligence and statistics}, pages 3457--3465. PMLR, 2021.

\bibitem[{Da~Veiga}(2015)]{da2015global}
Sébastien {Da~Veiga}.
\newblock Global sensitivity analysis with dependence measures.
\newblock \emph{Journal of Statistical Computation and Simulation}, 85\penalty0 (7):\penalty0 1283--1305, 2015.

\bibitem[{Da~Veiga} et~al.(2021){Da~Veiga}, Gamboa, Iooss, and Prieur]{DaVeiga2021}
Sébastien {Da~Veiga}, Fabrice Gamboa, Bertrand Iooss, and Clémentine Prieur.
\newblock \emph{Basics and {Trends} in {Sensitivity} {Analysis}: {Theory} and {Practice} in {R}}.
\newblock Society for Industrial and Applied Mathematics, Philadelphia, PA, 2021.
\newblock ISBN 978-1-61197-668-7 978-1-61197-669-4.
\newblock \doi{10.1137/1.9781611976694}.
\newblock URL \url{https://epubs.siam.org/doi/book/10.1137/1.9781611976694}.

\bibitem[Dai et~al.(2013)Dai, Ding, and Wahba]{dai2013multivariate}
Bin Dai, Shilin Ding, and Grace Wahba.
\newblock Multivariate bernoulli distribution.
\newblock \emph{Bernoulli}, 19\penalty0 (4):\penalty0 1465--1483, 2013.
\newblock \doi{10.3150/12-BEJSP10}.

\bibitem[Dua et~al.(2017)Dua, Graff, et~al.]{dua2017uci}
Dheeru Dua, Casey Graff, et~al.
\newblock Uci machine learning repository, 2017.
\newblock \emph{URL http://archive. ics. uci. edu/ml}, 7\penalty0 (1):\penalty0 62, 2017.

\bibitem[Gorji et~al.(2024)Gorji, Amrollahi, and Krause]{gorji2024shap}
Ali Gorji, Andisheh Amrollahi, and Andreas Krause.
\newblock Shap values via sparse {Fourier} representation.
\newblock \emph{Advances in Neural Information Processing Systems}, 2024.

\bibitem[Harsanyi(1963)]{harsanyi_1963}
John~C. Harsanyi.
\newblock A simplified bargaining model for the n-person cooperative game.
\newblock \emph{International Economic Review}, 4\penalty0 (2):\penalty0 194--220, 1963.
\newblock ISSN 00206598, 14682354.
\newblock URL \url{http://www.jstor.org/stable/2525487}.

\bibitem[Herin et~al.(2024)Herin, {Il~Idrissi}, Chabridon, and Iooss]{herin2024proportional}
Margot Herin, Marouane {Il~Idrissi}, Vincent Chabridon, and Bertrand Iooss.
\newblock Proportional marginal effects for global sensitivity analysis.
\newblock \emph{SIAM/ASA Journal on Uncertainty Quantification}, 12\penalty0 (2):\penalty0 667--692, 2024.

\bibitem[Herren and Hahn(2022)]{herren2022statistical}
Andrew Herren and P~Richard Hahn.
\newblock Statistical aspects of shap: Functional anova for model interpretation.
\newblock \emph{arXiv preprint arXiv:2208.09970}, 2022.

\bibitem[Hoeffding(1948)]{hoeffding_class_1948}
Wassily Hoeffding.
\newblock A class of statistics with asymptotically normal distribution.
\newblock \emph{Annals of Mathematical Statistics}, 19\penalty0 (3):\penalty0 293--325, 1948.
\newblock \doi{10.1214/aoms/1177730196}.
\newblock URL \url{https://projecteuclid.org/euclid.aoms/1177730196}.

\bibitem[Hooker(2007)]{hooker_2007}
Giles Hooker.
\newblock Generalized functional anova diagnostics for high-dimensional functions of dependent variables.
\newblock \emph{Journal of Computational and Graphical Statistics}, 16\penalty0 (3):\penalty0 709--732, 2007.
\newblock ISSN 10618600.
\newblock URL \url{http://www.jstor.org/stable/27594267}.

\bibitem[{Il~Idrissi} et~al.(2023){Il~Idrissi}, Bousquet, Gamboa, Iooss, and Loubes]{idrissi2023coalitional}
Marouane {Il~Idrissi}, Nicolas Bousquet, Fabrice Gamboa, Bertrand Iooss, and Jean-Michel Loubes.
\newblock On the coalitional decomposition of parameters of interest.
\newblock \emph{Comptes Rendus. Math{\'e}matique}, 361\penalty0 (G10):\penalty0 1653--1662, 2023.

\bibitem[{Il~Idrissi} et~al.(2025){Il~Idrissi}, Bousquet, Gamboa, Iooss, and Loubes]{IlIdrissi2023}
Marouane {Il~Idrissi}, Nicolas Bousquet, Fabrice Gamboa, Bertrand Iooss, and Jean-Michel Loubes.
\newblock Hoeffding decomposition of functions of random dependent variables.
\newblock \emph{Journal of Multivariate Analysis}, page 105444, 2025.

\bibitem[Koller and Friedman(2009)]{koller2009probabilistic}
Daphne Koller and Nir Friedman.
\newblock \emph{Probabilistic graphical models: principles and techniques}.
\newblock MIT press, 2009.

\bibitem[Labreuche(2022)]{LABREUCHE2022225}
Christophe Labreuche.
\newblock {Explanation with the Winter value: Efficient computation for hierarchical Choquet integrals}.
\newblock \emph{International Journal of Approximate Reasoning}, 151:\penalty0 225--250, 2022.

\bibitem[Lengerich et~al.(2020)Lengerich, Tan, Chang, Hooker, and Caruana]{lengerich2020purifying}
Benjamin Lengerich, Sarah Tan, Chun-Hao Chang, Giles Hooker, and Rich Caruana.
\newblock Purifying interaction effects with the functional anova: An efficient algorithm for recovering identifiable additive models.
\newblock In \emph{International Conference on Artificial Intelligence and Statistics}, pages 2402--2412. PMLR, 2020.

\bibitem[Lundberg and Lee(2017)]{lundberg2017unified}
Scott~M Lundberg and Su-In Lee.
\newblock A unified approach to interpreting model predictions.
\newblock \emph{Advances in neural information processing systems}, 30, 2017.

\bibitem[Lundberg et~al.(2018)Lundberg, Erion, and Lee]{lundberg2018consistent}
Scott~M Lundberg, Gabriel~G Erion, and Su-In Lee.
\newblock Consistent individualized feature attribution for tree ensembles.
\newblock \emph{arXiv preprint arXiv:1802.03888}, 2018.

\bibitem[Montgomery(2017)]{montgomery2017design}
Douglas~C Montgomery.
\newblock \emph{Design and analysis of experiments}.
\newblock John wiley \& sons, 2017.

\bibitem[O'Donnell(2014)]{o2014analysis}
Ryan O'Donnell.
\newblock \emph{Analysis of boolean functions}.
\newblock Cambridge University Press, 2014.

\bibitem[Owen and Prieur(2017)]{owen_shapley_2017}
Art~B Owen and Clémentine Prieur.
\newblock On shapley value for measuring importance of dependent inputs.
\newblock \emph{{SIAM}/{ASA} Journal on Uncertainty Quantification}, 5\penalty0 (1):\penalty0 986--1002, 2017.
\newblock ISSN 2166-2525.
\newblock \doi{10.1137/16M1097717}.
\newblock URL \url{https://epubs.siam.org/doi/10.1137/16M1097717}.

\bibitem[Poelwijk et~al.(2019)Poelwijk, Socolich, and Ranganathan]{poelwijk2019learning}
Frank~J Poelwijk, Michael Socolich, and Rama Ranganathan.
\newblock Learning the pattern of epistasis linking genotype and phenotype in a protein.
\newblock \emph{Nature communications}, 10\penalty0 (1):\penalty0 4213, 2019.

\bibitem[Rabitz and Ali{\c{s}}(1999)]{Rabitz1999}
Herschel Rabitz and {\"O}mer~Faruk Ali{\c{s}}.
\newblock General foundations of high-dimensional model representations.
\newblock \emph{Journal of Mathematical Chemistry}, 25\penalty0 (2):\penalty0 197--233, 1999.
\newblock ISSN 1572-8897.
\newblock \doi{10.1023/A:1019188517934}.
\newblock URL \url{https://doi.org/10.1023/A:1019188517934}.

\bibitem[Razavi et~al.(2021)Razavi, Jakeman, Saltelli, Prieur, Iooss, Borgonovo, Plischke, Lo~Piano, Iwanaga, Becker, Tarantola, Guillaume, Jakeman, Gupta, Melillo, Rabitti, Chabridon, Duan, Sun, Smith, Sheikholeslami, Hosseini, Asadzadeh, Puy, Kucherenko, and Maier]{Razavi2021}
Saman Razavi, Anthony~J Jakeman, Andrea Saltelli, Clémentine Prieur, Bertrand Iooss, Emanuele Borgonovo, Elmar Plischke, Samuele Lo~Piano, Takuya Iwanaga, William Becker, Stefano Tarantola, Joseph Guillaume, John Jakeman, Hoshin Gupta, Nicola Melillo, Giovanni Rabitti, Vincent Chabridon, Qingyun Duan, Xifu Sun, Stefán Smith, Razi Sheikholeslami, Nasim Hosseini, Masoud Asadzadeh, Arnald Puy, Sergei Kucherenko, and Holger Maier.
\newblock The future of {sensitivity} {analysis}: {An} essential discipline for systems modeling and policy support.
\newblock \emph{Environmental Modelling \& Software}, 137:\penalty0 104954, 2021.
\newblock ISSN 13648152.
\newblock \doi{10.1016/j.envsoft.2020.104954}.
\newblock URL \url{https://linkinghub.elsevier.com/retrieve/pii/S1364815220310112}.

\bibitem[Sarkisyan et~al.(2016)Sarkisyan, Bolotin, Meer, Usmanova, Mishin, Sharonov, Ivankov, Bozhanova, Baranov, Soylemez, et~al.]{sarkisyan2016local}
Karen~S Sarkisyan, Dmitry~A Bolotin, Margarita~V Meer, Dinara~R Usmanova, Alexander~S Mishin, George~V Sharonov, Dmitry~N Ivankov, Nina~G Bozhanova, Mikhail~S Baranov, Onuralp Soylemez, et~al.
\newblock Local fitness landscape of the green fluorescent protein.
\newblock \emph{Nature}, 533\penalty0 (7603):\penalty0 397--401, 2016.

\bibitem[Shapley(1951)]{shapley_notes_1951}
L.S. Shapley.
\newblock Notes on the n-person game -- {II}: The value of an n-person game, 1951.
\newblock URL \url{https://www.rand.org/content/dam/rand/pubs/research_memoranda/2008/RM670.pdf}.

\bibitem[Shrikumar et~al.(2017)Shrikumar, Greenside, and Kundaje]{shrikumar2017learning}
Avanti Shrikumar, Peyton Greenside, and Anshul Kundaje.
\newblock Learning important features through propagating activation differences.
\newblock In \emph{International conference on machine learning}, pages 3145--3153. PMlR, 2017.

\bibitem[Sobol'(1990)]{sobol1990sensitivity}
Il'ya~Meerovich Sobol'.
\newblock On sensitivity estimation for nonlinear mathematical models.
\newblock \emph{Matematicheskoe modelirovanie}, 2\penalty0 (1):\penalty0 112--118, 1990.

\bibitem[Song et~al.(2016)Song, Nelson, and Staum]{song2016shapley}
Eunhye Song, Barry~L Nelson, and Jeremy Staum.
\newblock {Shapley effects for global sensitivity analysis: Theory and computation}.
\newblock \emph{SIAM/ASA Journal on Uncertainty Quantification}, 4\penalty0 (1):\penalty0 1060--1083, 2016.

\bibitem[Stone(1994)]{stone1994use}
Charles~J Stone.
\newblock The use of polynomial splines and their tensor products in multivariate function estimation.
\newblock \emph{The annals of statistics}, pages 118--171, 1994.

\bibitem[Wu et~al.(2016)Wu, Dai, Olson, Lloyd-Smith, and Sun]{wu2016adaptation}
Nicholas~C Wu, Lei Dai, C~Anders Olson, James~O Lloyd-Smith, and Ren Sun.
\newblock Adaptation in protein fitness landscapes is facilitated by indirect paths.
\newblock \emph{Elife}, 5:\penalty0 e16965, 2016.

\bibitem[Yu et~al.(2019)Yu, Bien, and Tibshirani]{yu2019reluctant}
Guo Yu, Jacob Bien, and Ryan Tibshirani.
\newblock Reluctant interaction modeling.
\newblock \emph{arXiv preprint arXiv:1907.08414}, 2019.

\end{thebibliography}

\newpage

\onecolumn

\title{Fourier Analysis on the Boolean Hypercube \\ via Hoeffding Functional Decomposition\\ (Supplementary Material)}
\maketitle

\appendix

\section{ Experiment Details }

\subsection{Setup}

All the experiments were conducted on a standard personal laptop with 32 Go RAM. All experiments were done using Python and are displayed on standard Jupyter Notebooks. All computations are done on CPU.

\subsection{Data Processing}

Except the dataset \textbf{A} which is binary, all are categorical. They are binarized with standard one-hot encoding technique to work with only boolean features. Moreover, all (potential) constant columns are removed from the dataset and are assumed to be non significant. We recall the table of datasets displayed in the main paper:

\begin{table}[ht]
\caption{Benchmark Datasets. The task is either Classification (C) or Regression (R). $d$ denotes the feature dimension after data processing.}
\label{tab:datasets_bis}
\centering
\begin{tabular}{llcrr}
\toprule
ID & Dataset & Task & $d$ & $n$\\
\midrule
\textbf{A} & Entacmaea \citep{poelwijk2019learning} & R & 13 & 8\,192 \\
\textbf{B} & kr-vs-kp \citep{dua2017uci} & C & 35 & 3\,196 \\
\textbf{C} & SGEMM \citep{dua2017uci} & R & 40 & 241\,600 \\
\textbf{D} & GB1 \citep{wu2016adaptation} & R & 80 & 149\,361 \\
\textbf{E} & Mushrooms \citep{dua2017uci} & C & 116 & 8\,124 \\
\textbf{F} & avGFP \citep{sarkisyan2016local} & R & 233 & 54\,025 \\
\bottomrule
\end{tabular}
\end{table}

\subsection{Machine Learning Models}

We recall that we consider a \emph{black-box} setting. We train a machine learning model which gives a pseudo-Boolean function $f$ that we want to apply our framework on.

\paragraph{Random Forest.}
For the datasets \textbf{B} and \textbf{E}, we employed a scikit-learn Random Forest classifier. We utilized an ensemble of 100 estimators with a maximum depth of 5. Evaluation on the test set was reported using accuracy alongside a comprehensive classification report (precision, recall, and F1-score).

\paragraph{XGB.}
For the datasets \textbf{A-C-D-F}, we utilized XGBoost regressors optimized for the squared error objective across all these datasets. The evaluation metrics were strictly standardized to Root Mean Squared Error (RMSE) and $R^2$ scores on held-out test sets. Model hyperparameters were tailored as follows:

\begin{itemize}
    \item \textbf{Datasets \textbf{A} and \textbf{C}:} A baseline configuration was used, comprising 100 estimators, a learning rate of 0.1, and a maximum depth of 5.
    
    \item \textbf{Dataset \textbf{D}:} A deeper ensemble was required. We used 1,000 estimators with a depth of 10 and a learning rate of 0.05. Stochastic gradient boosting was introduced by setting both row and column subsampling to 80\%.
    
    \item \textbf{Dataset \textbf{F}:} To prevent overfitting on this specific set, we employed a highly regularized histogram-based XGBoost model. Hyperparameters included a learning rate of 0.01, L1/L2 regularization penalties (\texttt{alpha=0.1}, \texttt{lambda=1.0}), and strict constraints on leaf weights (\texttt{min\_child\_weight=5}). Furthermore, the training process was dynamically halted using early stopping with a 100-round patience based on validation set performance.
\end{itemize}

\paragraph{MLP.}
For the datasets \textbf{A-C-D-F}, we implemented Multi-Layer Perceptron (MLP) regressors using PyTorch, optimized for the Mean Squared Error (MSE) loss. Consistent with the tree-based models, predictive performance was evaluated on held-out test sets utilizing Root Mean Squared Error (RMSE) and $R^2$ scores. The neural network architectures and training procedures were adapted to each dataset's complexity as follows:

\begin{itemize}
    \item \textbf{Dataset \textbf{A}:} A standard feed-forward architecture was utilized, consisting of three hidden layers with 512, 256, and 64 units respectively, each followed by a ReLU activation. The model was trained over 100 epochs using the Adam optimizer with a learning rate of 0.001.
    
    \item \textbf{Dataset \textbf{C}:} We employed a network with hidden dimensions of 128, 64, and 32. To improve generalization, Batch Normalization and a Dropout rate of 20\% were applied to the first hidden layer. Training was conducted for 20 epochs using mini-batches of size 256.
    
    \item \textbf{Dataset \textbf{D}:} Following a regularization scheme similar to Dataset C, the architecture was widened to an initial hidden layer of 512 units (incorporating Batch Normalization and 20\% Dropout), followed by layers of 64 and 32 units. The network was trained over 10 epochs with a batch size of 256.
    
    \item \textbf{Dataset \textbf{F}:} To capture more intricate patterns while strictly preventing overfitting, a highly optimized architecture was deployed. The network featured wider layers (1024, 512, and 128 units), utilizing GELU activations, Batch Normalization across all hidden layers, and a 30\% Dropout rate on the first two. Optimization was performed using AdamW with weight decay ($10^{-4}$). Furthermore, a 90-10 train-validation split was introduced to dynamically decay the learning rate via a \texttt{ReduceLROnPlateau} scheduler (factor of 0.2, patience of 3), training with a larger batch size of 512 for 20 epochs.
\end{itemize}

\begin{table}[h]
    \centering
    \caption{Model performance across all datasets.}
    \label{tab:model_scores}
    \begin{tabular}{lcc}
        \toprule
        & \multicolumn{2}{c}{\textbf{Model Score}} \\
        \cmidrule(lr){2-3}
        \textbf{Dataset} & \textbf{Tree} & \textbf{MLP} \\
        \midrule
        \textbf{A} & 0.96 & 0.95 \\
        \textbf{B} & 0.93 & --   \\
        \textbf{C} & 0.94 & 0.99 \\
        \textbf{D} & 0.86 & 0.96 \\
        \textbf{E} & 0.99 & --   \\
        \textbf{F} & 0.59 & 0.72 \\
        \bottomrule
    \end{tabular}
\end{table}

\subsection{Empirical Fourier Analysis}
Recall that the family of functions to conduct our \emph{generalized} Fourier analysis is given by:
\begin{equation}
    \forall S \subseteq [d], \forall \mathbf x \in \mathcal{X}, \psi_S( \mathbf x ) = \frac{ (-1)^{ \sum\limits_{i \in S} \mathbf x_i } }{ 2^{ \vert S \vert } \cdot p_S( \mathbf x_S) }.
\end{equation}
In practice, we worked with sets $S$ such that $\vert S \vert \leq 2$ to bound the number of functions to $\mathcal{O}(d^2)$. We minimized the following Elastic Net cost function:
\begin{equation}
\min\limits_{ \bm \beta \in \mathbb{R}^{r} } \frac{1}{2n} \| \bm y_n - \bm \Psi_{red} \bm \beta \|_{ \mathbb{R}^n }^2 + \alpha \left( \lambda \| \bm \beta \|_1 + \frac{1-\lambda}{2} \| \bm \beta \|_2^2 \right).
\end{equation}
The optimization was performed using the scikit-learn framework, where the parameters $\alpha$ and $\lambda$ correspond to the \texttt{alpha} and \texttt{l1\_ratio} arguments, respectively. Furthermore, the underlying coordinate descent algorithm was constrained by a maximum number of iterations, denoted as \texttt{max\_iter}. We also denote by \texttt{matrix\_shape} the shape of the design matrix $\bm{\Psi}_{red}$ for each approximation of order $k$. We display all these details in the following table:

\begin{table}[ht]
\caption{Table of hyperparameters taken for each dataset.}
\label{tab:hyperparam}
\centering
\begin{tabular}{cccccc}
\toprule
ID & $k$ & \texttt{matrix\_shape} & \texttt{alpha} & \texttt{l1\_ratio} & \texttt{max\_iter}\\
\midrule
\textbf{A} & 2 & $\left( 8\,192, 92 \right)$ & 1e-4 & 0.5 & 5\,000 \\
\textbf{B} & 1 & $\left( 2\,891, 36 \right)$ & 1e-4 & 0.5 & 5\,000 \\
\textbf{C} & 2 & $\left( 241\,600, 821 \right)$ & 1e-1 & 0.5 & 5\,000 \\
\textbf{D} & 2 & $\left( 149\,361, 3241 \right)$  & 1e-2 & 0.5 & 5\,000 \\
\textbf{E} & 1 & $\left( 8\,124, 117 \right)$  & 1e-2 & 0.5 & 5\,000 \\
\textbf{F} & 1 & $\left( 49\,089, 234 \right)$  & 1e-4 & 0.5 & 5\,000 \\
\bottomrule
\end{tabular}
\end{table}

For each dataset, we also display the computation time to obtain the entire decomposition. We run the framework which first compute the matrix $\bm{\Psi}_{red}$ and then solves the least squares problem for 6 times : one \emph{warm up} and five \emph{effective} runs. All the runs where performed on Jupyter notebook cells and we report the result in the following table:

\begin{table}[ht]
\caption{Table of computation time over 5 effective runs on Jupyter notebook cells.}
\label{tab:time}
\centering
    \begin{tabular}{lcc}
        \toprule
        & \multicolumn{2}{c}{Mean Time (s) $\pm$ std (s)} \\
        \cmidrule(lr){2-3}
        \textbf{Dataset} & \textbf{Tree} & \textbf{MLP} \\
        \midrule
        \textbf{A} & $0.034036 \pm 0.007250$ & $0.048663 \pm 0.011523$ \\
        \textbf{B} & $0.023216 \pm 0.002210$ & --   \\
        \textbf{C} & $34.840913 \pm 3.740652$ & $36.704091 \pm 2.210165$ \\
        \textbf{D} & $117.657666 \pm 8.828244$ & $121.720641 \pm 12.719668$ \\
        \textbf{E} & $0.184458 \pm 0.026737$ & --   \\
        \textbf{F} & $0.652419 \pm 0.021759$ & $0.647476 \pm 0.030388$ \\
        \bottomrule
    \end{tabular}
\end{table}
Note that once the computation is done, we have access to the entire decomposition on the dataset, which provides global analysis and explainability.

\subsection{SHAP}

In the standard \texttt{shap} \citep{lundberg2017unified} package, the mathematical quantities computed by KernelSHAP and TreeSHAP differ in both their theoretical target and their computational approach. Let $d$ be the input dimension, $S \subseteq [d]$ a coalition of present features, and $\mathbf{x}_S$ the observed feature values for a given instance $\mathbf{x}$.

\paragraph{KernelSHAP.}
KernelSHAP \citep{lundberg2017unified} computes an approximation of the interventional Shapley values. Given a model $f$ and a background dataset $\mathbf{X}_{bg}$ of size $B$, the value function $v(S)$ is defined as the marginal expectation over the omitted features $\bar{S}$:
\begin{equation}
v(S) = \mathbb{E}_{\mathbf{X}_{bg}} [f(\mathbf{x}_S, \mathbf{X}_{\bar{S}})] \approx \frac{1}{B} \sum_{b=1}^B f(\mathbf{x}_S, \mathbf{x}_{\bar{S}}^{(b)}),
\end{equation}
where $\mathbf{x}_{\bar{S}}^{(b)}$ is drawn from the background distribution. To avoid the $\mathcal{O}(2^d)$ combinatorial complexity, KernelSHAP randomly samples coalitions and estimates the feature attributions $\phi_i$ by solving a weighted least squares regression problem:
\begin{equation}
\min_{\phi_0, \dots, \phi_d} \sum_{S \subseteq \{1, \dots, d\}} \pi(S) \left( v(S) - \left( \phi_0 + \sum_{i \in S} \phi_i \right) \right)^2,
\end{equation}
where the weighting function $\pi(S) = \frac{d-1}{\binom{d}{|S|}|S|(d-|S|)}$ is the Shapley kernel (see \citet{covert2021improving} for further details).

\paragraph{DeepSHAP.}
For deep neural network architectures, such as the Multi-Layer Perceptrons (MLPs) employed in our study, the \texttt{shap} package utilizes DeepSHAP. DeepSHAP is a high-speed approximation algorithm that builds upon the DeepLIFT framework \citep{shrikumar2017learning} to estimate Shapley values. Instead of explicitly evaluating feature coalitions like KernelSHAP, it leverages the network's internal gradient structure. Given a background dataset $\mathbf{X}_{bg}$, DeepSHAP attributes the difference between the target prediction $f(\mathbf{x})$ and the expected baseline prediction to the $d$ input features:
\begin{equation}
f(\mathbf{x}) - \mathbb{E}_{\mathbf{X}_{bg}} [f(\mathbf{X})] \approx \sum_{i=1}^d \phi_i.
\end{equation}
It achieves this by computing ``multipliers'' (discrete gradients) that backpropagate the activation differences from the output layer down to the input features via a modified chain rule. By averaging these DeepLIFT attributions over multiple reference samples drawn from $\mathbf{X}_{bg}$, DeepSHAP efficiently approximates the interventional Shapley values $v(S)$ in a single backward pass per reference, making it highly scalable for deep architectures.

\paragraph{TreeSHAP.}
Conversely, TreeSHAP \citep{lundberg2018consistent} operates directly on the internal structure of tree-based ensembles, and the quantity it computes strictly depends on the initialization of the \texttt{TreeExplainer}. By default, when no background dataset is provided, TreeSHAP computes the exact \textit{path-dependent} (conditional) expectation:
\begin{equation}
v(S) = \mathbb{E} [f(\mathbf{X}) \mid \mathbf{X}_S = \mathbf{x}_S].
\end{equation}
It achieves this analytically by tracing decision paths and weighting node predictions by their corresponding training sample coverage (the proportion of training samples passing through each node). The fundamental advantage is that TreeSHAP leverages the trees' \emph{recursive structure} to compute the \emph{exact} Shapley values in polynomial time ( $\mathcal{O}(T L D^2)$, where $T$ is the number of trees, $L$ the maximum number of leaves, and $D$ the maximum depth).

\paragraph{Experiment details.} For our attribution experiments, we utilized the standard implementations of KernelSHAP, DeepSHAP, and TreeSHAP provided by the \texttt{shap} library. Table \ref{tab:shap_configs} details the configuration for each method across the evaluated datasets. Specifically, $n_{\text{inst}}$ denotes the number of instances selected for explanation, where ``All'' indicates that the method was applied to the entire dataset. The parameter $n_{\text{bg}}$ represents the size of the background dataset, which is exclusively required by the Kernel and Deep explainers. As reflected in the table, TreeSHAP was applied to the tree ensembles across all six datasets, whereas KernelSHAP and DeepSHAP were specifically deployed to explain the Multi-Layer Perceptrons trained on the four regression datasets (\textbf{A}, \textbf{C}, \textbf{D}, and \textbf{F}).

\begin{table}[ht]
    \centering
    \caption{Summary of SHAP configurations: number of explained instances ($n_{\text{inst}}$) and background sizes ($n_{\text{bg}}$) across datasets.}
    \label{tab:shap_configs}
    \begin{tabular}{lcccccc}
        \toprule
        & \multicolumn{2}{c}{\textbf{TreeSHAP}} & \multicolumn{2}{c}{\textbf{KernelSHAP}} & \multicolumn{2}{c}{\textbf{DeepSHAP}} \\
        \cmidrule(lr){2-3} \cmidrule(lr){4-5} \cmidrule(lr){6-7}
        \textbf{Dataset} & $n_{\text{inst}}$ & $n_{\text{bg}}$ & $n_{\text{inst}}$ & $n_{\text{bg}}$ & $n_{\text{inst}}$ & $n_{\text{bg}}$ \\
        \midrule
        \textbf{A} & All & -- & 1 & 500  & 1 & 500 \\
        \textbf{B} & All & -- & -- & -- & -- & --  \\
        \textbf{C} & All & -- & 1 & 500 & 1 & 500 \\
        \textbf{D} & All & -- & 1 & 500 & 1 & 500 \\
        \textbf{E} & 2\,000 & -- & -- & -- & -- & -- \\
        \textbf{F} & 2\,000 & -- & 1 & 500 & 1 & 500 \\
        \bottomrule
    \end{tabular}
\end{table}

\paragraph{SHAP Based Indicators.}
To assess global feature importance, we define distinct normalized explainability indicators depending on the attribution method. For TreeSHAP, we evaluate the overall magnitude of the contributions by aggregating the absolute Shapley values over all instances:
\begin{equation}
    I_i^{\text{tree}} \coloneqq \frac{\sum_{\mathbf{x} } \vert \phi_i( \mathbf{x}) \vert}{ \sum_{j=1}^d \sum_{\mathbf{x} } \vert \phi_j( \mathbf{x}) \vert }.
\end{equation}
Conversely, for KernelSHAP and DeepSHAP, we measure the signed impact by considering the raw Shapley values. To maintain standardized values, this net contribution is subsequently normalized by the $\ell_1$ norm of the aggregated scores across all $d$ features:
\begin{equation}
    I_i^{\text{kernel/deep}} \coloneqq \frac{\phi_i(\mathbf{x})}{ \sum_{j=1}^d \left\vert  \phi_j(\mathbf{x}) \right\vert }.
\end{equation}
We display these indicators on the barplots of the main paper.

\paragraph{Our Indicators.}
Recall that the HFD of $f$ is given by:
\begin{equation}
    f(\mathbf x) = \sum\limits_{S \subseteq [d]} f_S( \mathbf x_S ).
\end{equation}
For a given order $k$, we consider the following reduced approximation of $f$:
\begin{equation}
    f_{red}^{k}( \mathbf x ) \coloneqq \sum\limits_{\vert S \vert \leq k} f_S( \mathbf x_S ).
\end{equation}

A natural local explanation indicator can be defined as follows:
\begin{equation}
    I_{i}^{\text{Fourier}} \coloneqq \frac{ \sum_{S \ni i} f_S(\mathbf x_S) { \vert S \vert }^{-1} }{ \sum_{j=1}^{d} \left\vert \sum_{S \ni j} f_S(\mathbf x_S) { \vert S \vert }^{-1} \right\vert },
\end{equation}
which can be seen as a $k-$order truncated (and normalized) local Shapley value. The natural associated global explanation indicator is simply given by the $\ell_1$ norm of the local (non-normalized) Shapley value that we normalize at the end.

Across datasets and model classes, our method produces feature rankings that are qualitatively consistent with these baselines, both in terms of the induced ordering and the relative magnitudes of the dominant effects. 
This agreement provides an empirical validation of our measure-dependent functional decomposition as a statistically grounded reference quantity, and supports the interpretation of widely used SHAP variants as approximations of comparable global and local effects under an implicit choice of distribution.
Beyond agreement, our approach also offers a computational advantage: once the coefficients of the truncated expansion are estimated, local and global attributions can be obtained by simple evaluations and aggregations, yielding low marginal cost per explained instance.

\subsection{Supplementary Details on Entacmaea Dataset}

The dataset \text{A} is particularly interesting because the empirical measure is exactly uniform on the Boolean hypercube. Indeed, there are exactly 13 binary features for $2^{13}$ rows in the dataset. So the empirical measure is given by:
\begin{equation}
   \forall \mathbf x \in \{0,1\}^{13}, \: p( \mathbf x) = \frac{1}{2^{13}}.
\end{equation}
In this case, the design matrix $\bm \Psi $ shall be orthogonal and the corresponding Gram matrix $G$ shall be diagonal. Furthermore, the feature attribution given by our method should particularly coincide with SHAP.

Indeed, we compute the Frobenius norm between $G$ and the diagonal matrix of diagonal coefficients of $G$ and we obtain as expected a near $0$ distance : $ 1.62e-18 $.

\begin{figure*}[t!] 
    \centering

    \begin{subfigure}[b]{\gridscale\dimexpr0.32\textwidth\relax}
        \centering
        \includegraphics[width=\linewidth]{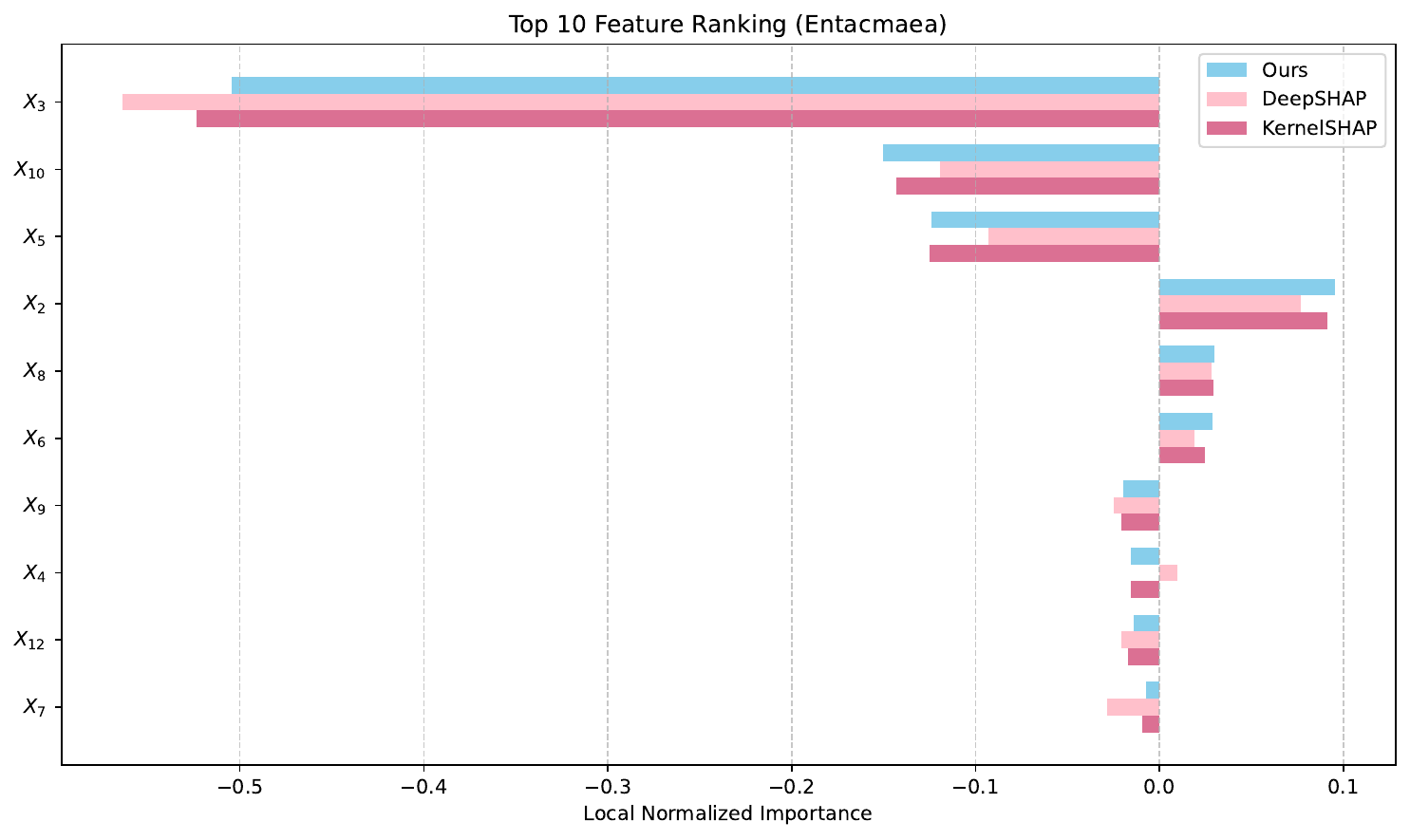}
        \caption{Entacmaea (1)}
        \label{subfig:entacmaea_random_1}
    \end{subfigure}
    \hfill
    \begin{subfigure}[b]{\gridscale\dimexpr0.32\textwidth\relax}
        \centering
        \includegraphics[width=\linewidth]{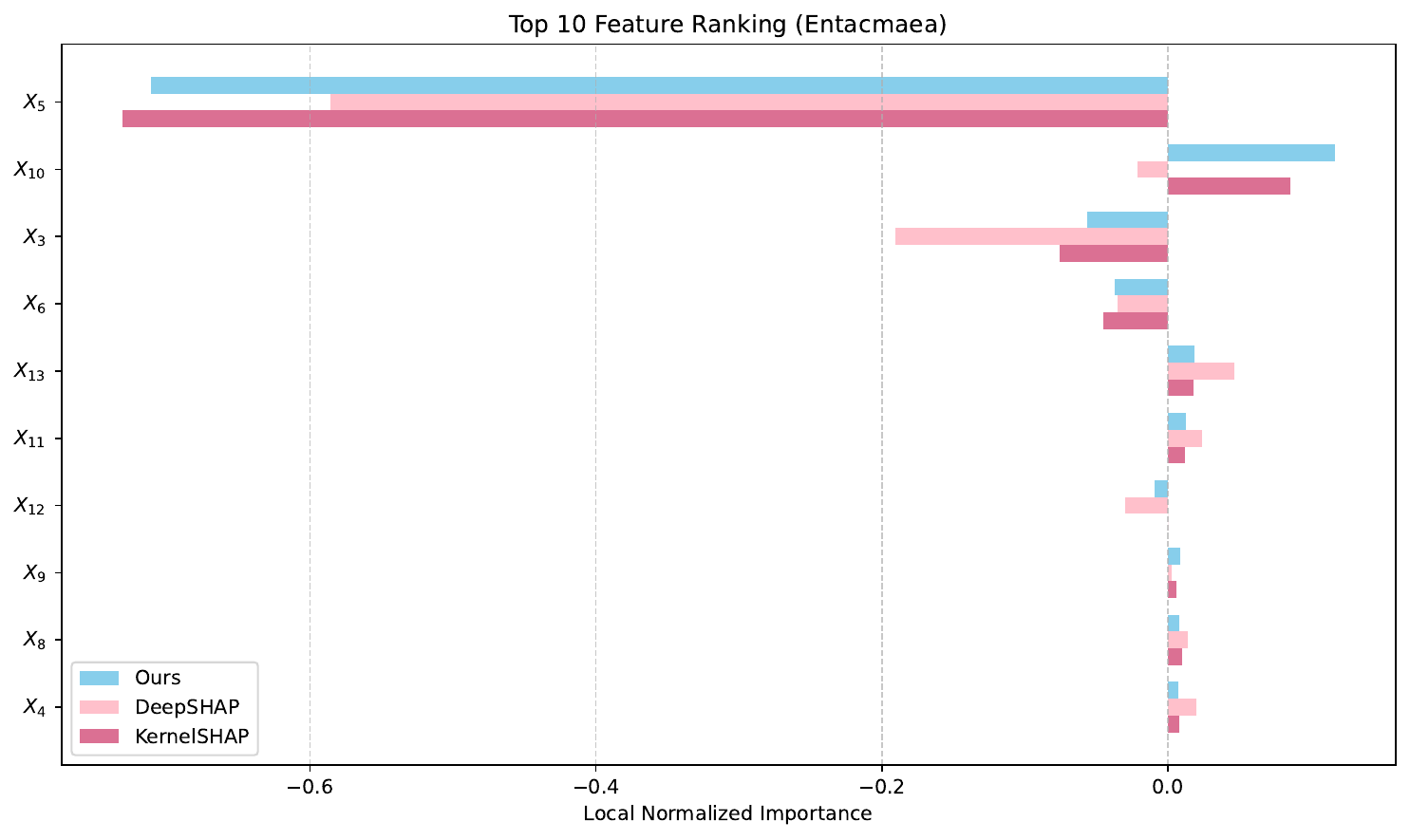}
        \caption{Entacmaea (2)}
        \label{subfig:entacmaea_random_2}
    \end{subfigure}
    \hfill
    \begin{subfigure}[b]{\gridscale\dimexpr0.32\textwidth\relax}
        \centering
        \includegraphics[width=\linewidth]{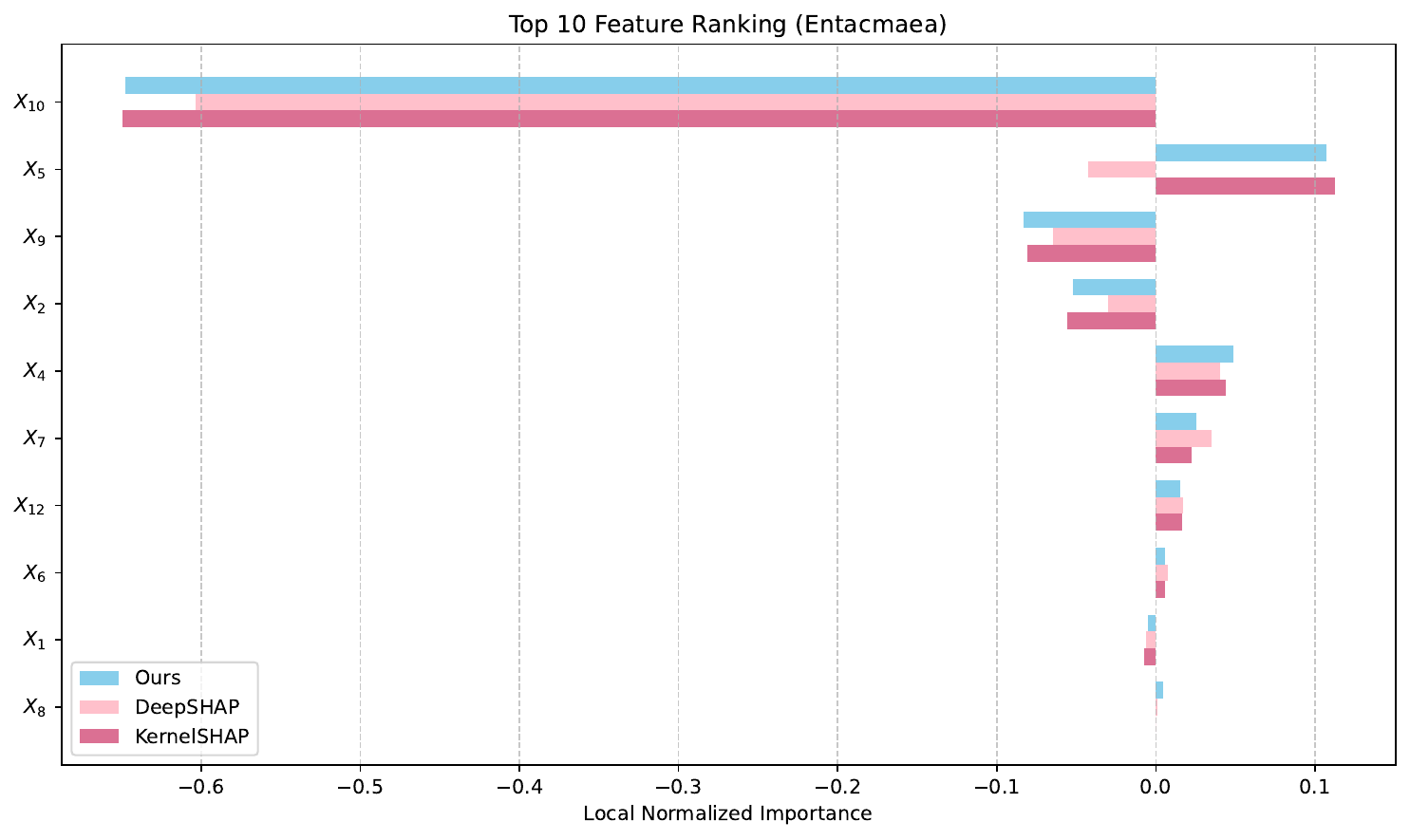}
        \caption{Entacmaea (3)}
        \label{subfig:entacmaea_random_3}
    \end{subfigure}

    \vspace{0.35cm} 

    \begin{subfigure}[b]{\gridscale\dimexpr0.32\textwidth\relax}
        \centering
        \includegraphics[width=\linewidth]{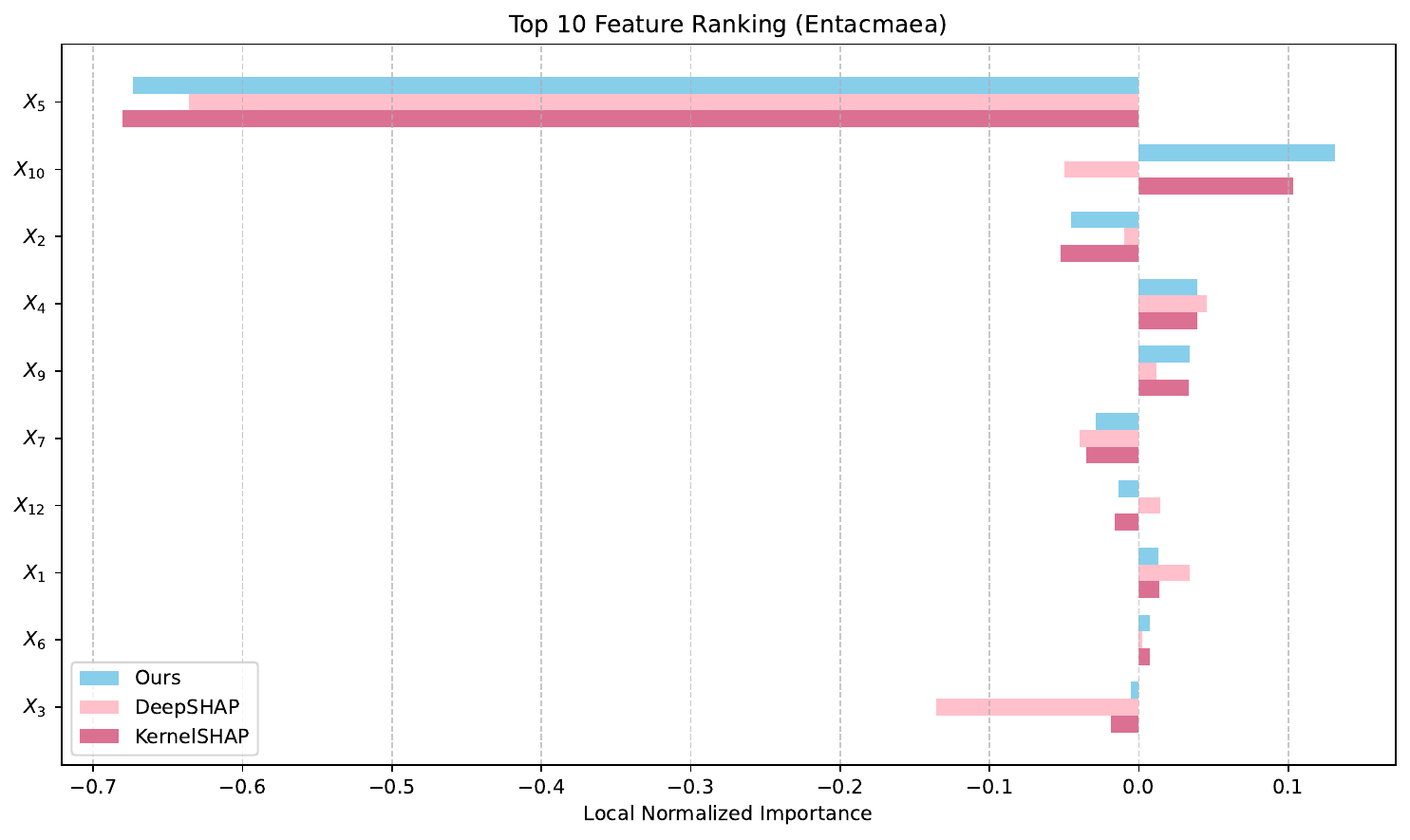}
        \caption{Entacmaea (4)}
        \label{subfig:entacmaea_random_4}
    \end{subfigure}
    \hfill
    \begin{subfigure}[b]{\gridscale\dimexpr0.32\textwidth\relax}
        \centering
        \includegraphics[width=\linewidth]{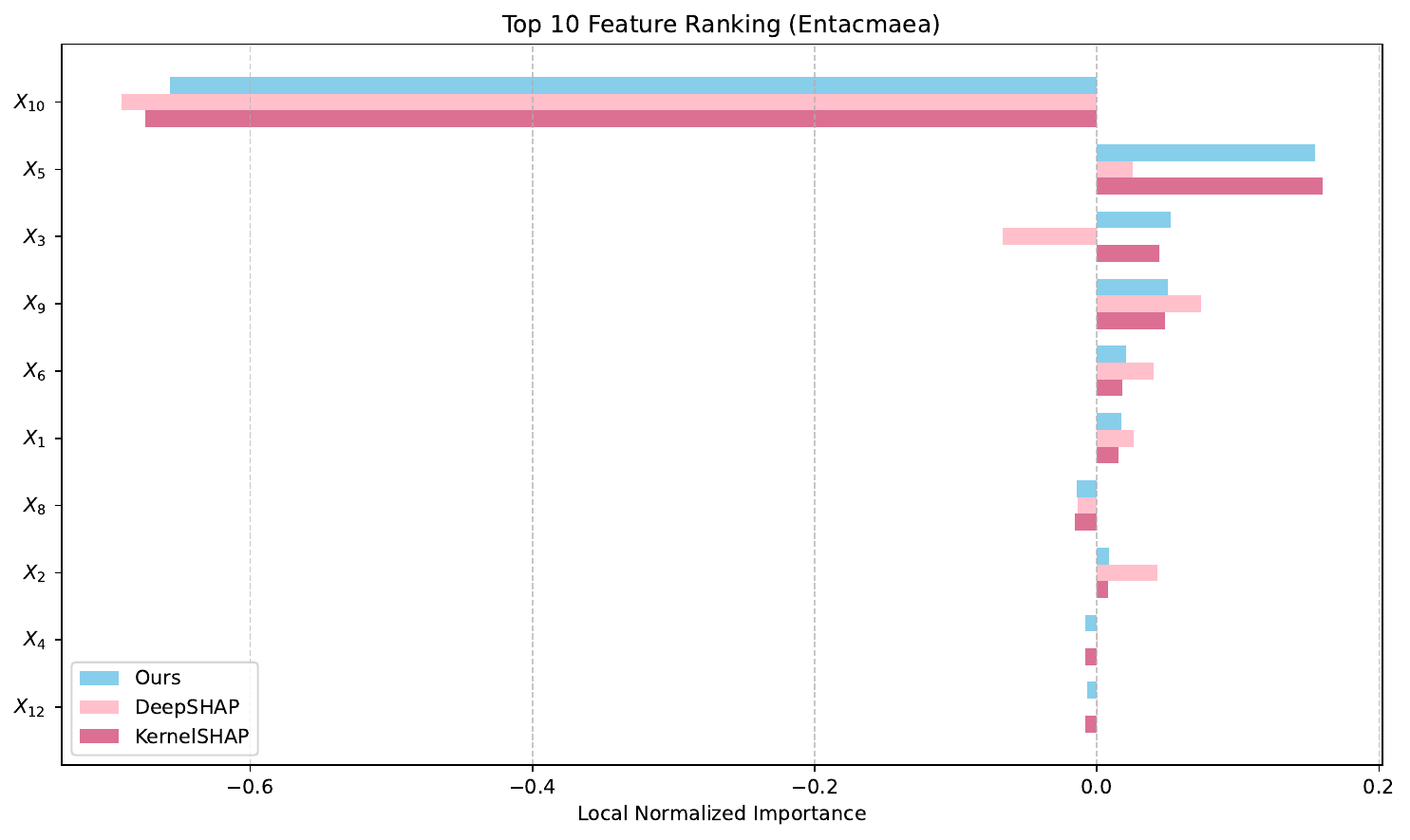}
        \caption{Entacmaea (5)}
        \label{subfig:entacmaea_random_5}
    \end{subfigure}
    \hfill
    \begin{subfigure}[b]{\gridscale\dimexpr0.32\textwidth\relax}
        \centering
        \includegraphics[width=\linewidth]{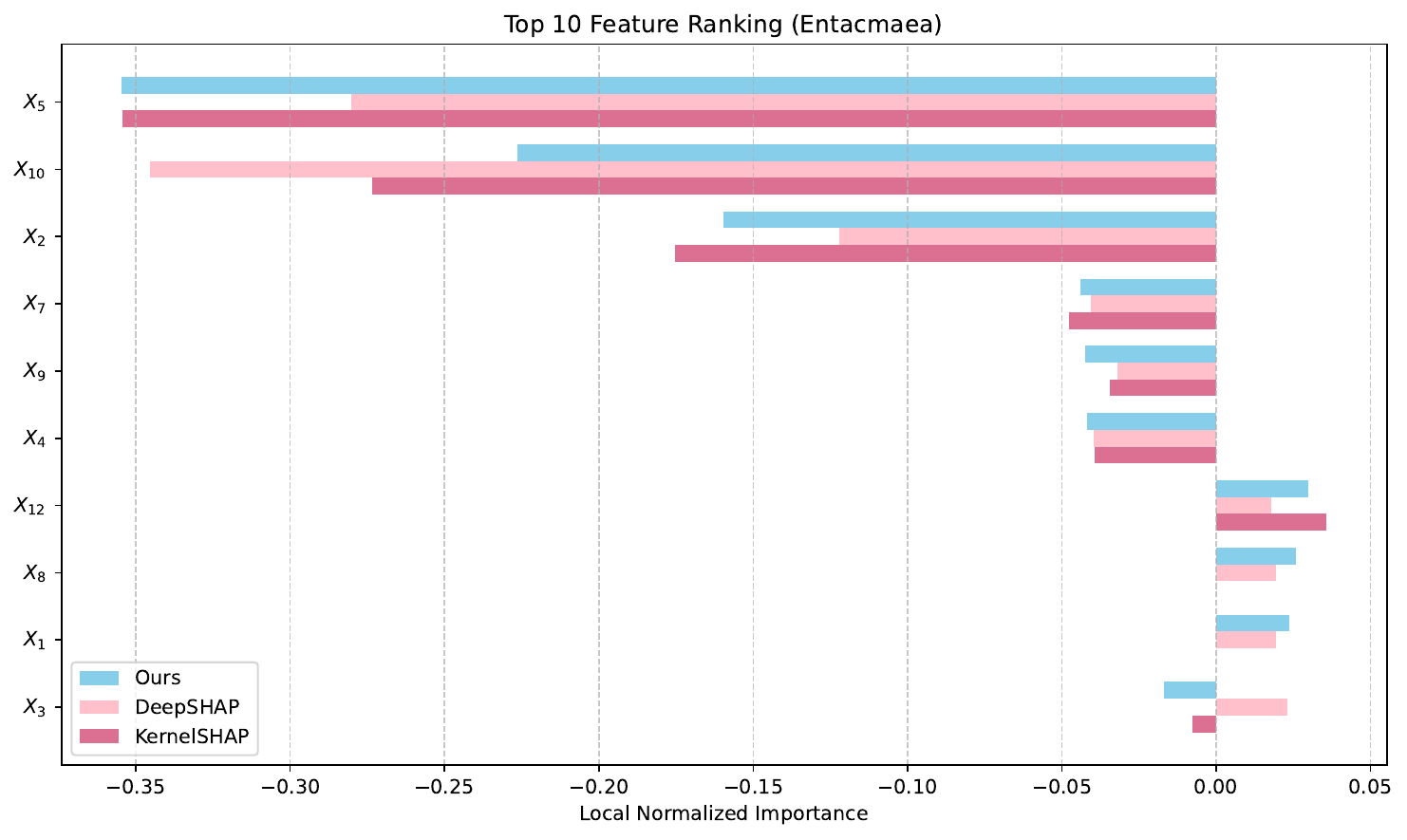}
        \caption{Entacmaea (6)}
        \label{subfig:entacmaea_random_6}
    \end{subfigure}

    \vspace{0.35cm} 

    \begin{subfigure}[b]{\gridscale\dimexpr0.32\textwidth\relax}
        \centering
        \includegraphics[width=\linewidth]{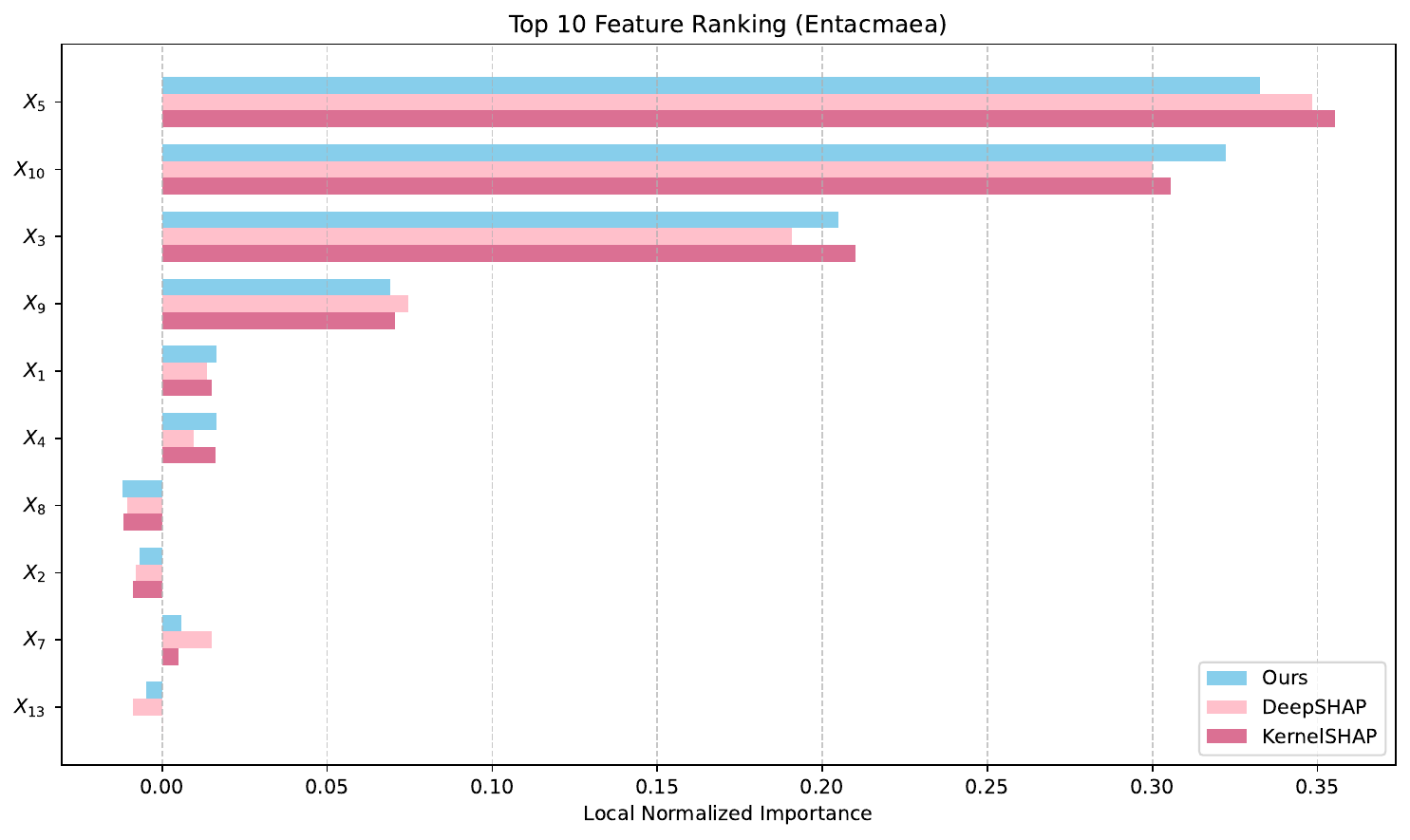}
        \caption{Entacmaea (7)}
        \label{subfig:entacmaea_random_7}
    \end{subfigure}
    \hfill
    \begin{subfigure}[b]{\gridscale\dimexpr0.32\textwidth\relax}
        \centering
        \includegraphics[width=\linewidth]{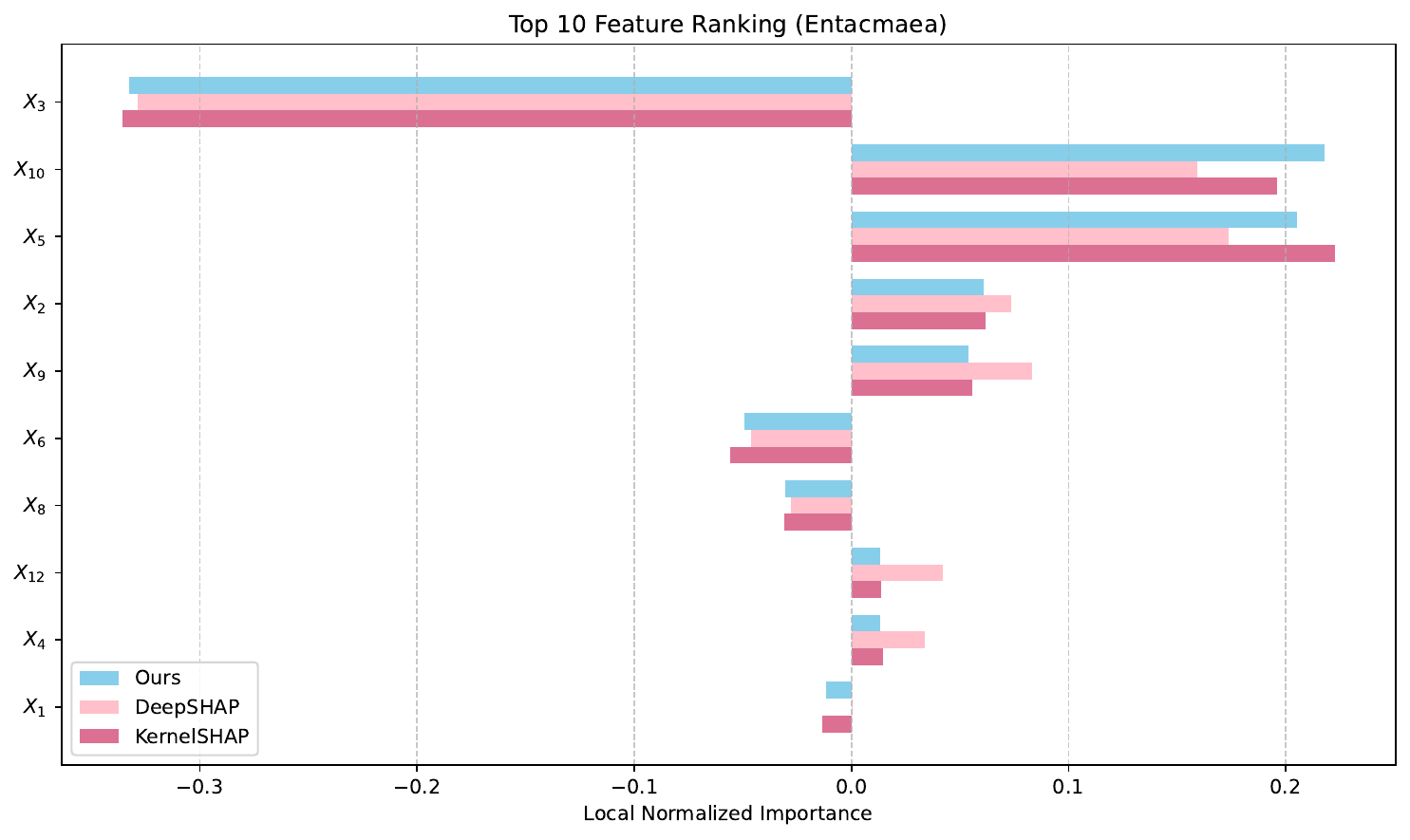}
        \caption{Entacmaea (8)}
        \label{subfig:entacmaea_random_8}
    \end{subfigure}
    \hfill
    \begin{subfigure}[b]{\gridscale\dimexpr0.32\textwidth\relax}
        \centering
        \includegraphics[width=\linewidth]{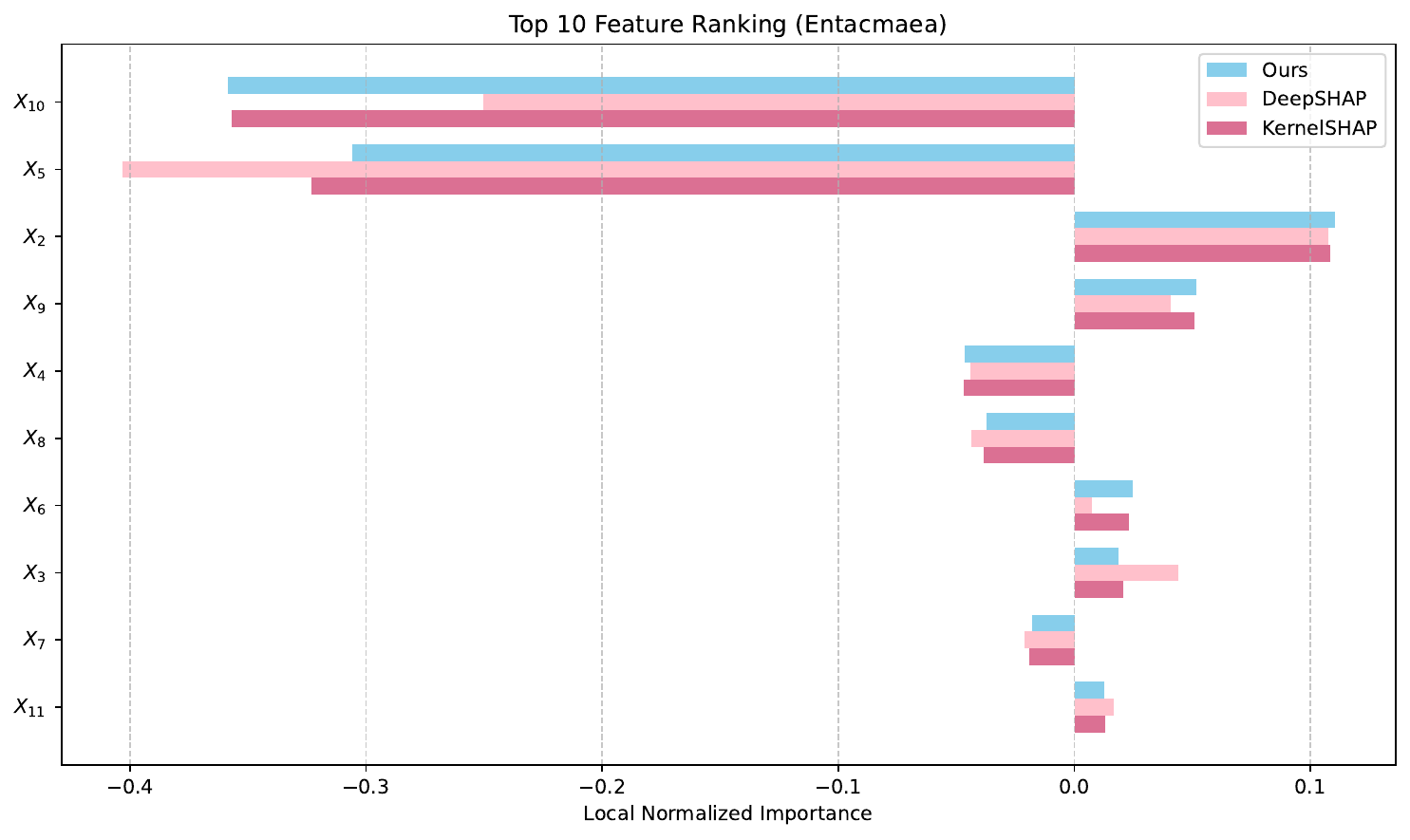}
        \caption{Entacmaea (9)}
        \label{subfig:entacmaea_random_9}
    \end{subfigure}

    \vspace{0.35cm} 

    \begin{subfigure}[b]{\gridscale\dimexpr0.32\textwidth\relax}
        \centering
        \includegraphics[width=\linewidth]{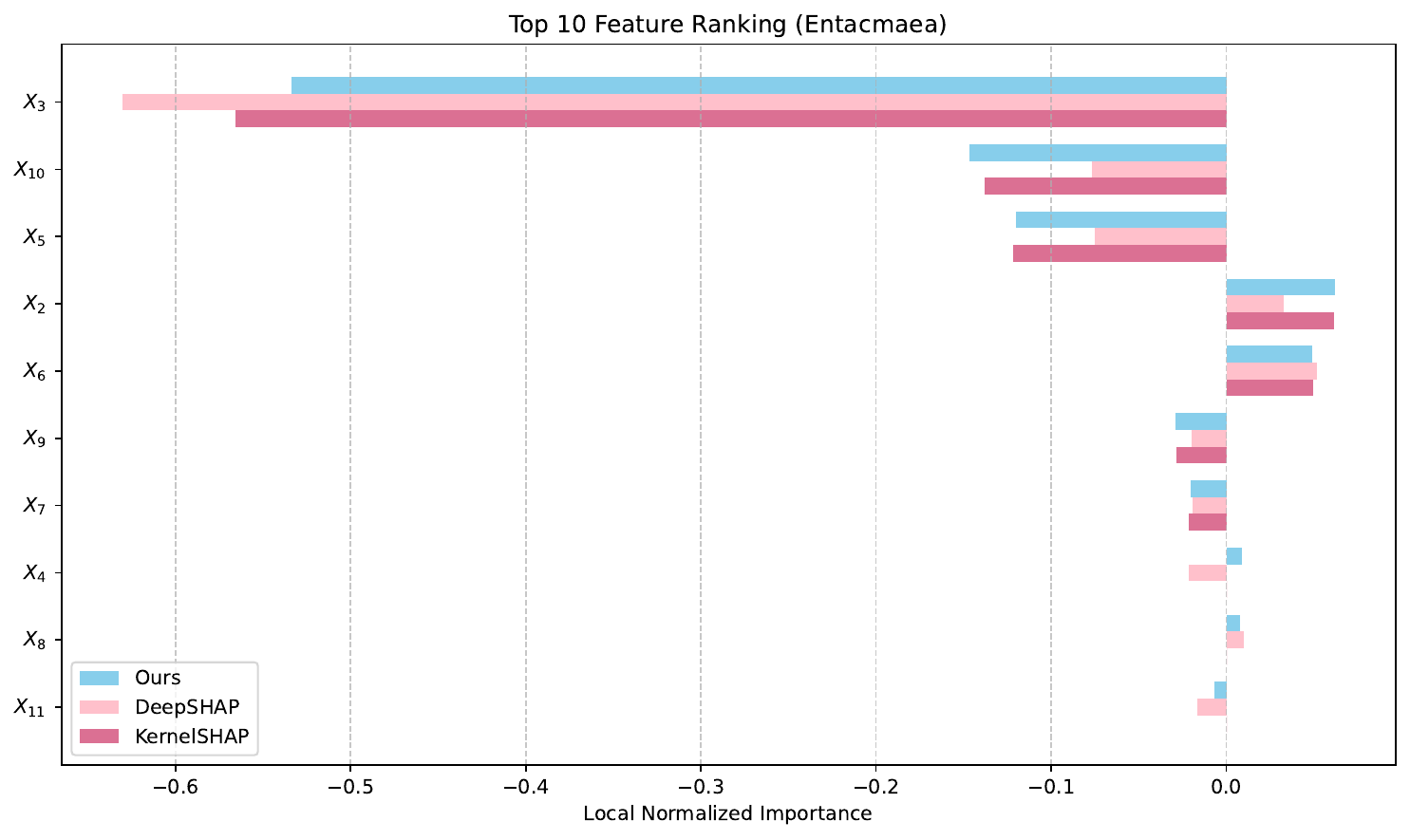}
        \caption{Entacmaea (10)}
        \label{subfig:entacmaea_random_10}
    \end{subfigure}
    \hfill
    \begin{subfigure}[b]{\gridscale\dimexpr0.32\textwidth\relax}
        \centering
        \includegraphics[width=\linewidth]{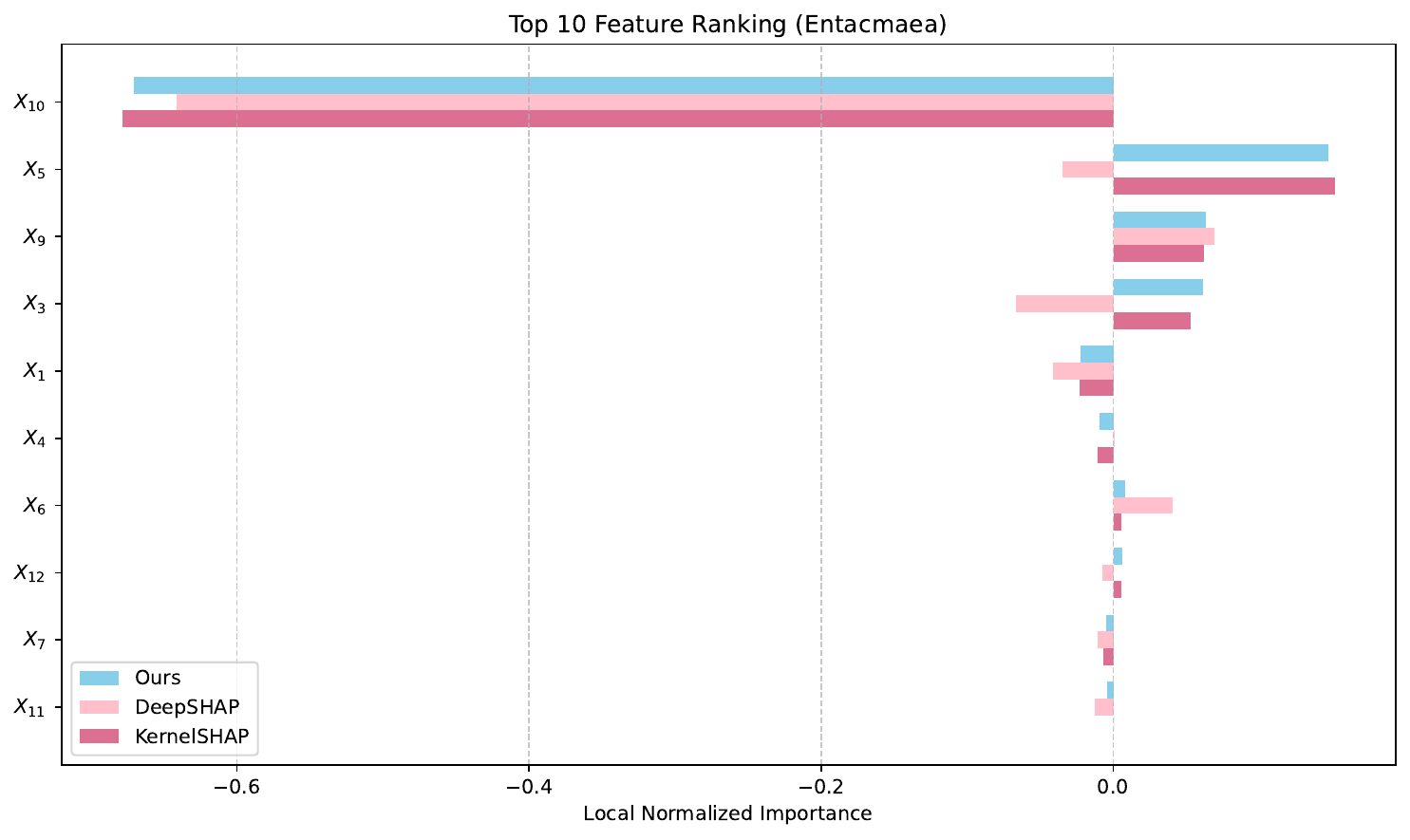}
        \caption{Entacmaea (11)}
        \label{subfig:entacmaea_random_11}
    \end{subfigure}
    \hfill
    \begin{subfigure}[b]{\gridscale\dimexpr0.32\textwidth\relax}
        \centering
        \includegraphics[width=\linewidth]{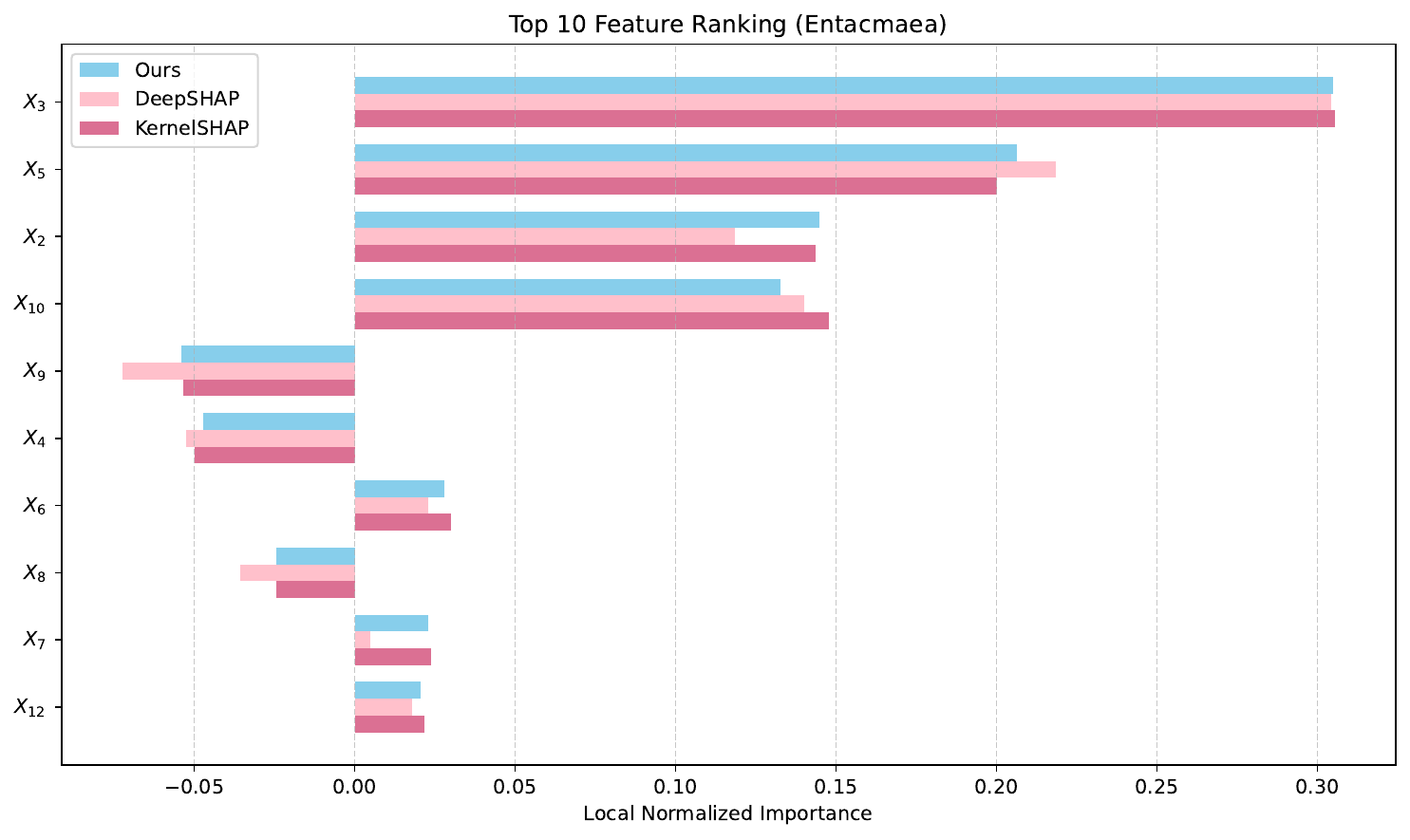}
        \caption{Entacmaea (12)}
        \label{subfig:entacmaea_random_12}
    \end{subfigure}

    \caption{\textbf{Global feature importance on MLP for Dataset \textbf{A} (Entacmaea).} For each random run, we report the top-10 features ranked by global importance according to our method, comparing KernelSHAP and DeepSHAP to our method. Across all twelve runs, the induced attributions are consistent: the rankings and values are very close.}
    \label{fig:supp_entacmaea}
\end{figure*}

We also compare on 12 different instances sampled at random the feature attribution given by our method and by SHAP and we observe near equal importances. We report our results in Fig. \ref{fig:supp_entacmaea}.

\newpage
\section{ Illustration of Correlations }

In this section, we provide geometric details concerning Fig. \ref{fig:GX_dependence_levels}. Let us consider 2 random variables $\mathbf X_1 , \mathbf X_2 \in \{0,1\}^2$ such that:
\begin{eqnarray}
    \mathbb{P}\left( \mathbf X_1 = 0 , \mathbf X_2 = 0 \right) = \mathbb{P}\left( \mathbf X_1 = 1 , \mathbf X_2 = 1 \right) &=& q, \\
    \mathbb{P}\left( \mathbf X_1 = 0 , \mathbf X_2 = 1 \right) = \mathbb{P}\left( \mathbf X_1 = 1 , \mathbf X_2 = 0 \right) &=& \frac{1}{2} - q,
\end{eqnarray}
where $q \in [0 , \frac{1}{2}]$ is a parameter that controls the dependence between $\mathbf X_1$ and $\mathbf X_2$, in this case $\mathbf X_1$ and $\mathbf X_2$ are i.i.d. iff $q = \frac{1}{4}$. In this toy study case, the dimension $d$ equals 2 and the corresponding Boolean hypercube is a square with 4 vertices for the 4 possible configurations. The basis function $\psi_1$ and $\psi_2$ are simply given by:
\begin{eqnarray}
    \psi_1( \mathbf X_1 ) = (-1)^{\mathbf X_1}, \\
    \psi_2( \mathbf X_2 ) = (-1)^{ \mathbf X _2 }.
\end{eqnarray}
First, one can compute the correlation $\mathrm{corr}$ between $\mathbf X_1$ and $\mathbf X_2$ which is given by:
\begin{equation}
    \mathrm{corr} = 4q - 1.
\end{equation}
Then we can compute the scalar product between $\psi_1$ and $\psi_2$ which is:
\begin{equation}
    \left\langle \: \psi_1 \: , \: \psi_2 \: \right\rangle = \mathbb{E}\left[ \psi_1( \mathbf X_1 ) \psi_2( \mathbf X_2 ) \right] = 4q - 1 = \mathrm{corr}.
\end{equation}
Using that $\psi_1$ and $\psi_2$ are normalized, this scalar product can be seen also as the cosine of the angle $\theta$ between $\psi_1$ and $\psi_2$ in the usual Euclidean plan, which finally gives:
\begin{equation}
    \theta = \arccos(\mathrm{corr}),
\end{equation}
and justifies the illustration given in Fig. \ref{fig:GX_dependence_levels}.

\section{ Proofs }

\begin{lemma}
    Under any distribution $\mathbb P$ on the Hilbert space $\mathcal H$, for any subset $S \subseteq [d]$, $\psi_S( \mathbf X )$ is a square-integrable function of $\mathbf X_S$.
\end{lemma}
\begin{proof}
    Recall that $\psi_S = \frac{ \chi_S( \mathbf X ) }{ 2^{ \vert S \vert } \cdot p_S( \mathbf X_S )  }$. This is clearly a measurable function of $\mathbf X_S$. Furthermore, this function is always defined since $p_S( \mathbf X_S) \neq 0$, indeed, $\mathbf X_S$ takes values in its own support and since it is a discrete random vector its pmf never equals $0$ on its support. Finally, since $\mathbf X_S$ takes a finite number of values, $\psi_S( \mathbf X )$ takes also a finite number of real values and so is a bounded random variable.
\end{proof}

\subsection{ Full Support Assumption }

Under this hypothesis, recall that the HFD exists and is unique. The corresponding Hilbert space $\mathcal H$ can be decomposed as follows:
\begin{equation}
    \mathcal{H} = \bigoplus\limits_{S \subseteq [d]} V_S,
\end{equation}
the spaces $V_S$ are called the \emph{Hoeffding decomposition spaces}. They satisfy the following recursion (see \citet{IlIdrissi2023}):
\begin{itemize}
    \item $V_{\emptyset}$ is the space of $\mathbb P-$almost surely constant functions
    \item $V_{S}$ is the space of functions of $\mathbf X_S$ such that for all $T \subsetneq S$, the elements of $V_S$ are orthogonal to the elements of $V_T$
\end{itemize}
More formally, the recursion is defined as follows:
\begin{equation}
    \forall S \subseteq [d], \: V_S \coloneqq \left( \bigoplus\limits_{T \subsetneq S} V_T \right)^{\perp_S},
\end{equation}
where the symbol $\perp_S$ refers to the orthogonality complement in the space of functions of $\mathbf X_S$.
This construction ensures that finding a suitable functional basis $v \coloneqq \left\{ v_S \right\}_{S \subseteq [d]}$ such that for all $S$ $v_S$ span $V_S$ solves the optimization program formulated by Hooker. To prove our main result (\textit{i.e.} Theorem \ref{thm:gen_fourier}) under the full support assumption we will show that:
\begin{itemize}
    \item these subsets are one dimensional ; 
    \item for any subset $S$, $V_S$ is spanned by $\psi_S$.
\end{itemize}
These two points will suffice to prove 
Theorem \ref{thm:gen_fourier} because it is 
exactly equivalent to say that:
\begin{itemize}
    \item $\left\{ \psi_S \right\}_{S \subseteq [d]}$ is a basis of $\mathcal H$ which ensures the existence and the uniqueness of the Fourier expansion ;
    \item the corresponding decomposition satisfies ANOVA formulation \eqref{eq:def_anova} and hierarchical orthogonality \eqref{eq:orthogonality}.
\end{itemize}

\begin{proposition}
    Under the hypothesis of full support, the family of random variables $\left( \psi_S( \mathbf X ) \right)_{S \subseteq [d]}$ satisfies the \underline{hierarchical orthogonality condition}, \textit{i.e.}
    \begin{equation}
        \forall T \subsetneq S \subseteq [d], \: \mathbb{E}\left[ \psi_S( \mathbf X ) \psi_T( \mathbf X) \right] = 0.
    \end{equation}
\end{proposition}
\begin{proof}
    Let $T \subsetneq S \subseteq [d]$, we have that both $\psi_S( \mathbf X)$ and $\psi_T(\mathbf X)$ are measurable functions of $\mathbf X_S$. The expectation can be computed under the distribution of $\mathbf X_S$ which gives:
    \begin{equation}
        \mathbb{E}\left[ \psi_S( \mathbf X ) \psi_T( \mathbf X) \right] = \sum\limits_{ \mathbf{z} \in \{0,1\}^{ \vert S \vert } } \psi_S( \mathbf z) \psi_T( \mathbf z) p_S( \mathbf z ).
    \end{equation}
    We denote $\mathcal X_S$ the support of $\mathbf X_S$ and by $\mathbf z_T$ the components of $\mathbf z$ indexed by $T$. This previous equation can be written as follows:
    \begin{equation}
        \mathbb{E}\left[ \psi_S( \mathbf X ) \psi_T( \mathbf X) \right] = \sum\limits_{ \mathbf{z} \in \{0,1\}^{ \vert S \vert } } \frac{ \chi_S( \mathbf z ) }{\textcolor{orange}{2^{ \vert S \vert }} \cdot \textcolor{red}{p_S( \mathbf z )}} \cdot \frac{ \chi_T( \mathbf z ) }{\textcolor{orange}{2^{ \vert T \vert }} \cdot p_T( \mathbf z_T )} \cdot \textcolor{red}{p_S( \mathbf z)},
    \end{equation}
    which simplifies as:
    \begin{equation}
        \mathbb{E}\left[ \psi_S( \mathbf X ) \psi_T( \mathbf X) \right] \propto \sum\limits_{ \mathbf z \in \{0,1\}^{ \vert S \vert } } \frac{ \textcolor{blue}{(-1)^{ \sum\limits_{a \in T} \mathbf z_a} } \cdot (-1)^{ \sum\limits_{b \notin T} \mathbf z_b } \cdot \textcolor{blue}{(-1)^{ \sum\limits_{c \in T} \mathbf z_c} } }{ p_T( \mathbf z_T ) },
    \end{equation}
    finally giving:
    \begin{equation}
        \mathbb{E}\left[ \psi_S( \mathbf X ) \psi_T( \mathbf X) \right] \propto \sum\limits_{ \mathbf z \in \{0,1\}^{ \vert S \vert } } \frac{ (-1)^{ \sum\limits_{i \in S \setminus T} \mathbf z_i } }{ p_T( \mathbf z_T ) }.
    \end{equation}
    The previous sum can be written as follows:
    \begin{equation}
        \mathbb{E}\left[ \psi_S( \mathbf X ) \psi_T( \mathbf X) \right] \propto \sum\limits_{ \mathbf v \in \{0,1\}^{ \vert T \vert } } \frac{1}{p_T( \mathbf v )} \cdot \sum\limits_{ \mathbf{u} \in \{0,1\}^{ \vert S \vert } \: : \: \mathbf{u}_T = \mathbf v } (-1)^{ \sum\limits_{ i \in S \setminus T } \mathbf u_i }.
    \end{equation}
    When we fix a configuration vector $\mathbf v \in \{0,1\}^{ \vert T \vert }$, the sum $ \sum\limits_{ \mathbf{u} \in \{0,1\}^{ \vert S \vert } \: : \: \mathbf{u}_T = \mathbf v } (-1)^{ \sum\limits_{ i \in S \setminus T } \mathbf u_i } $ is over $2^{ \vert S \vert - \vert T \vert }$ to \emph{complete} the configuration $\mathbf v$ and equals $0$.
\end{proof}

\begin{proposition}
Under the full support assumption, for all $S \subseteq [d]$, the Hoeffding decomposition space $V_S$ is one-dimensional and is given by:
\begin{equation}
    V_S = \mathrm{span}\left( \psi_S(\mathbf X_S) \right).
\end{equation}
\end{proposition}

\begin{proof}
We proceed by induction on the cardinality of $S \subseteq [d]$. Let $\mathcal{H}_S$ denote the space of pseudo-Boolean functions depending only on the variables in $\mathbf X_S$. Under the full support assumption, the dimension of this space is strictly $\dim(\mathcal{H}_S) = 2^{|S|}$.

\begin{itemize}
    \item {\it Base case ($|S| = 0$)}: The space $\mathcal{H}_{\emptyset}$ is the space of constant functions, so $\dim(\mathcal{H}_{\emptyset}) = 2^0 = 1$. By definition, $V_{\emptyset} = \mathcal{H}_{\emptyset}$. Since $\psi_{\emptyset}(\mathbf X_{\emptyset}) = 1$ almost surely, it is a non-zero constant, which directly implies $V_{\emptyset} = \mathrm{span}(\psi_{\emptyset})$.
    
    \item {\it Inductive step}: Let $c \in \mathbb{N}$ and assume that for all $T \subseteq [d]$ such that $|T| \leq c$, we have $V_T = \mathrm{span}(\psi_T)$. Now let $S \subseteq [d]$ be a subset with $|S| = c+1$. By the definition of the Hoeffding decomposition, the space $\mathcal{H}_S$ can be written as the direct sum of the spaces $V_T$ for $T \subseteq S$:
    \begin{equation}
        \mathcal{H}_S = \bigoplus_{T \subseteq S} V_T = \left( \bigoplus_{T \subsetneq S} V_T \right) \oplus V_S.
    \end{equation}
    By the inductive hypothesis, for all $T \subsetneq S$, $\dim(V_T) = 1$. Therefore, the dimension of the direct sum of these strictly smaller spaces is exactly the number of strict subsets of $S$:
    \begin{equation}
        \dim \left( \bigoplus_{T \subsetneq S} V_T \right) = \sum_{T \subsetneq S} 1 = 2^{|S|} - 1.
    \end{equation}
    Since $\dim(\mathcal{H}_S) = 2^{|S|}$, it immediately follows that $\dim(V_S) = 1$.
    
    Now, we must show that $\psi_S(\mathbf X_S)$ spans this one-dimensional space $V_S$. First, $\psi_S$ is clearly a function of $\mathcal{H}_S$. Furthermore, by the hierarchical orthogonality established in the previous proposition, $\psi_S$ is orthogonal to $\psi_T$ for all $T \subsetneq S$. Thanks to the inductive hypothesis, $V_T = \mathrm{span}(\psi_T)$, which means $\psi_S$ is orthogonal to the entirety of $V_T$ for all $T \subsetneq S$. 
    
    Consequently, $\psi_S \in \left( \bigoplus_{T \subsetneq S} V_T \right)^{\perp_S}$, which is precisely the definition of $V_S$. Since $\psi_S$ is a non-zero function belonging to the one-dimensional space $V_S$, it forms a basis for it, yielding $V_S = \mathrm{span}(\psi_S)$.
\end{itemize}
This concludes the induction.
\end{proof}
\begin{remark}
    First, note that this result holds only in the very particular case of full support hypothesis on the space of pseudo-Boolean functions. These subspaces are no longer one-dimensional if we consider continuous or discrete (with more than 2 modalities) settings. Furthermore, there is an \emph{exact} bijection between the number of configurations in this space and the number of subsets of HFD because of the full support assumption.
\end{remark}

\begin{tcolorbox}[blue_style]
The direct implication of this last proposition is the following result:
\begin{equation}
    \mathcal H = \bigoplus\limits_{ S \subseteq [d] } \mathrm{span}\left( \psi_S \right).
\end{equation}
This decomposition of $\mathcal H$ is the algebraic representation of Theorem \ref{thm:gen_fourier}.
\end{tcolorbox}
\begin{remark}
    A nice consequence of this result is the so-called \underline{exclusion property}. Indeed, the decomposition is unique, so for any pseudo Boolean function $f$, if there exists a strict subset $S \subseteq [d]$ such that $f( \mathbf X)$ is only a function of $\mathbf X_S$, then the set of non zero Fourier coefficients of $f$ in the generalized HFD representation of $f$ is given by:
    \begin{equation}
        \left\{\widehat{f}( T) \: \mid \: T \subseteq S \right\}.
    \end{equation}
    This means, in particular, that input variables that play no causal role in the prediction are automatically eliminated by the feature importance methods proposed by our approach.
\end{remark}

\begin{theorem}
    In the particular case where $\mathbb P$ is a product measure on $\mathcal H$, \textit{i.e.} if the components of $\mathbf X$ are mutually independent, the basis $\left\{ \psi_S \right\}_{ S \subseteq [d] }$ is orthogonal. The corresponding decomposition recovers the standard HFD \eqref{eq:anova_indep}.
\end{theorem}

\begin{proof}
    Let $S \neq T \subseteq [d]$, we will show that:
    \begin{equation}
        \left\langle \: \psi_S \: , \: \psi_T \: \right\rangle = 0.
    \end{equation}
    We have:
    \begin{align}
        \left\langle \: \psi_S \: , \: \psi_T \: \right\rangle &= \mathbb{E}\left[ \psi_S( \mathbf X ) \psi_T( \mathbf X ) \right], \\
        &= \mathbb E \left[ \frac{ \chi_S( \mathbf X) \cdot \chi_T(\mathbf X) }{ 2^{ \vert S\vert + \vert T \vert } \cdot p_S( \mathbf X_S) \cdot p_T( \mathbf X_T ) } \right], \\
        &\propto \mathbb{E}\left[ \prod\limits_{ s \in S }\left( \frac{ (-1)^{ \mathbf X_s } }{ p_s( \mathbf X_s) } \right) \cdot \prod\limits_{ t \in T }\left( \frac{ (-1)^{ \mathbf X_t } }{ p_t( \mathbf X_t) } \right) \right], \\
        &\propto \mathbb E \left[ \left( \prod\limits_{a \in S \cap T} \frac{ (-1)^{ \mathbf X_a } }{ p_a( \mathbf X_a ) } \right)^2 \cdot \prod\limits_{ b \in (S \cup T) \setminus ( S \cap T ) } \frac{ (-1)^{ \mathbf X_b } }{ p_b( \mathbf X_b ) } \right], \\
        &\propto \underbrace{\mathbb{E}\left[ \prod\limits_{a \in S \cap T} \frac{ 1 }{ p_a( \mathbf X_a )^2 } \right]}_{ >0 } \mathbb E \left[ \prod\limits_{ b \in (S \cup T) \setminus ( S \cap T ) } \frac{ (-1)^{ \mathbf X_b } }{ p_b( \mathbf X_b ) } \right], \\
        &\propto \prod\limits_{ b \in (S \cup T) \setminus ( S \cap T ) } \mathbb{E} \left[ \frac{ (-1)^{ \mathbf X_b } }{ p_b( \mathbf X_b ) } \right], \\
        &\propto \mathbb{E} \left[ \frac{ (-1)^{ \mathbf X_i } }{ p_i( \mathbf X_i ) } \right] \quad \text{for any chosen $i \in (S \cup T) \setminus ( S \cap T ) $}, \\
        &\propto \frac{ 1 }{ p_i(0) } p_i(0) + \frac{ -1 }{ p_i(1) } p_i(1), \\
        &= 0.
    \end{align}
   This concludes on the mutual orthogonality of the basis $\left\{ \psi_S \right\}_{ S \subseteq [d] }$.
\end{proof}

\subsection{Non Full Support}

In the non full support case, the collection of proposed functions $\left\{ \psi_S\right\}_{S \subseteq [d]}$ lost the basis property. However, we will show that it keeps generating the entire Hilbert space $\mathcal H$. Furthermore, one has to admit that the HFD is not well defined anymore. Indeed, since the dimension of $\mathcal H$ is no longer $2^d$, we lose the \emph{direct-sum} decomposition and so the uniqueness. The only thing that we can say is that we can extract \emph{a} representation of a pseudo-Boolean function $f$ from the generative family of functions $\left\{ \psi_S \right\}_{S \subseteq [d]}$.

\begin{lemma}
    Under any distribution $\mathbb P$ (and so any support $\mathcal X$) the family of functions $\left\{ \chi_S \right\}_{S \subseteq [d]}$ linearly spans $\mathcal H$.
\end{lemma}

\begin{proof}
    Let $f$ be a pseudo Boolean function defined on $\mathcal H$, we \emph{extend} $f$ by defining the following function:
    \begin{equation}
      \forall \mathbf x \in \{0,1\}^d, \:  f_{\text{full}} (\mathbf x) \coloneqq f( \mathbf x) \cdot \mathbf{1}_{ \left\{ \mathbf x \in \mathcal X \right\} },
    \end{equation}
    $f_{\text{full}}$ is an element of the Hilbert space of pseudo Boolean functions endowed with uniform measure. Consequently, $f_{\text{full}}$ admits a Fourier expansion on the entire Boolean hypercube as follows:
    \begin{equation}
        \forall \mathbf x \in \{0,1\}^d, \: f_{\text{full}}( \mathbf x) = \sum\limits_{S \subseteq [d]} c_S \cdot \chi_S( \mathbf x),
    \end{equation}
    which can be restricted as follows on the support $\mathcal X$:
    \begin{equation}
        f = \sum\limits_{S \subseteq [d]} c_S \cdot \chi_S.
    \end{equation}
    Finally, the family of parity functions still spans $\mathcal H$ equipped with any probability measure $\mathbb P$ but loses its orthogonality and its basis property.
\end{proof}

\begin{proposition}
    Under any distribution $\mathbb P$ (and so any support $\mathcal X$), the family of functions $\left\{ \psi_S \right\}_{S \subseteq [d]}$ linearly spans $\mathcal H$.
\end{proposition}

\begin{proof}
    By the previous lemma, the family of parity functions $\{ \chi_S \}_{S \subseteq [d]}$ spans $\mathcal H$. For any $S \subseteq [d]$, we define the weight function $W_S$ depending only on $\mathbf X_S$ as:
    \begin{equation}
        W_S(\mathbf X_S) \coloneqq \frac{1}{2^{\vert S \vert} \cdot p_S(\mathbf X_S)}.
    \end{equation}
    Since $p_S(\mathbf X_S) > 0$ almost surely under $\mathbb P$, $W_S$ is a well-defined function that depends strictly on the variables in $S$. Therefore, its standard Fourier expansion only involves subsets $T \subseteq S$:
    \begin{equation}
        W_S(\mathbf X_S) = \sum_{T \subseteq S} \omega_T^{(S)} \cdot \chi_T(\mathbf X).
    \end{equation}
    By definition, we have $\psi_S = W_S \cdot \chi_S$. Furthermore, we recall this nice property of the parity functions:
    \begin{equation}
        \forall T \subseteq S \subseteq[d], \: \chi_S \cdot \chi_T = \chi_{ S \setminus T },
    \end{equation}
    we then obtain:
    \begin{equation}
        \psi_S = \sum_{T \subseteq S} \omega_T^{(S)} \cdot \chi_{S \setminus T}.
    \end{equation}
    We can rewrite this sum as follows:
    \begin{equation}
        \psi_S = \sum\limits_{R \subseteq S} \omega_{S \setminus R}^{(S)} \cdot \chi_R.
    \end{equation}
    By denoting $v_{\psi}$ the vector of $[\psi_{\emptyset}, \dots, \psi_{[d]}]^{\top}$ and $v_{\chi}$ the vector of $[ \chi_{\emptyset}, \dots, \chi_{[d]} ]^{\top}$, one has:
    \begin{equation}
        v_{\psi} = W v_{\chi},
    \end{equation}
    where $W$ is the triangular matrix defined as:
    \begin{equation}
        \forall S, R \subseteq [d], \: W_{S,R} \coloneqq \omega_{ S \setminus R }^{(S)} \cdot \mathbf{1}_{ \left\{ R \subseteq S \right\} }.
    \end{equation}
    Finally, diagonal coefficients of $W$ are given by:
    \begin{eqnarray}
        W_{S,S} &=& \omega_{\emptyset}^{(S)} \\
                &=& \mathbb{E}_{ \text{unif} } \left[ W_S( \mathbf X_S ) \right] \\
                &=& \sum\limits_{ \mathbf x \in\{0,1\}^d } \frac{1}{2^d} \cdot W_S( \mathbf x_S) \cdot \mathbf{1}_{ \left\{ \mathbf x \in \mathcal X \right\} } \\
                &=& \frac{1}{2^{ d + \vert S \vert }} \sum\limits_{ \mathbf x \in \mathcal X } \frac{1}{p_S( \mathbf x_S)}
    \end{eqnarray}
    Since the diagonal coefficients of $W$ are all positive and using that $W$ is triangular, $W$ is invertible and one has:
    \begin{equation}
        v_{ \chi } = W^{-1} v_{\psi},
    \end{equation}
    consequently each parity function is spanned by the collection of functions $\left\{ \psi_S \right\}_{S \subseteq [d]}$. Since the parity functions span the entire Hilbert space $\mathcal H$, we have the desired result.
\end{proof}

\end{document}